\documentclass[twoside,11pt]{article}
\usepackage[all]{xy}
\usepackage{amsfonts,amsmath,amsthm}
\usepackage{color}
\usepackage{graphicx}
\usepackage{stmaryrd}        % provides \llbracket and \rrbracket
\usepackage{algorithm,algpseudocode}
\usepackage{subfigure}
\usepackage{cite}

\usepackage[paper=a4paper,dvips,top=2cm,left=2cm,right=2cm,
    foot=1cm,bottom=4cm]{geometry}

\newtheorem{theorem}{Theorem}[section]
\newtheorem{definition}{Definition}
\newtheorem{lemma}[theorem]{Lemma}

\def\H{\mathcal{H}}
\def\Y{\mathcal{Y}}
\def\R{\mathbb{R}}

\def\X{\mathcal{X}}
\def\O{\mathcal{O}}
\def\G{\mathcal{G}}

\def\N{\mathcal{N}}
\def\L{\mathfrak{L}}

\def\St{\textrm{St}}
\def\W{\mathcal{W}}

\begin{document}

\title{2D+3D facial expression recognition via embedded tensor manifold regularization}
\author{ Yunfang Fu\footnote{%
    School of Computer Science and Engineering, Shijiazhuang University, Shijiazhuang 050035, China, Institute of Information Science, Beijing Jiaotong University, Beijing 100044, China,  and Beijing Key Laboratory of Advanced Information Science and Network Technology, Beijing 100044, China.  e-mail: fu$\_$yunfang@126.com.}
     \and  \
    Qiuqi Ruan\footnote{%
    Institute of Information Science, Beijing Jiaotong University, Beijing 100044, China, and Beijing Key Laboratory of Advanced Information Science and Network Technology, Beijing 100044, China.  e-mail:  qqruan@bjtu.edu.cn.}
    \and  \
    Ziyan Luo\footnote{%
    Corresponding author, Department of Mathematics,
  Beijing Jiaotong University, Beijing 100044, China. ({\tt zyluo@bjtu.edu.cn}). }
    \and  \
    Gaoyun An\footnote{%
    Institute of Information Science, Beijing Jiaotong University, Beijing 100044, China, and Beijing Key Laboratory of Advanced Information Science and Network Technology, Beijing 100044, China.  e-mail: gyan@bjtu.edu.cn.}
    \and \
    Yi Jin\footnote{%
    Institute of Information Science, Beijing Jiaotong University, Beijing 100044, China, and Beijing Key Laboratory of Advanced Information Science and Network Technology, Beijing 100044, China.  e-mail: yjin@bjtu.edu.cn.}
    \and \
    Jun Wan\footnote{%
    National Laboratory of Pattern Recognition, Institute of Automation, Chinese Academy of Sciences, Beijing 100190, China. e-mail: jun.wan@nlpr.ia.ac.cn.}
}
\date{\today}
\maketitle

\begin{abstract}
In this paper, a novel approach via embedded tensor manifold regularization for 2D+3D facial expression recognition (FERETMR) is proposed.  Firstly, 3D tensors are constructed from 2D face images and 3D face shape models to keep the structural information and correlations.
To maintain the local structure (geometric information) of 3D tensor samples in the low-dimensional tensors space during the dimensionality reduction, the $\ell_0$-norm of the core tensors and a tensor manifold regularization scheme embedded on core tensors are adopted via a low-rank truncated Tucker decomposition on the generated tensors.
As a result, the obtained factor matrices will be used for facial expression classification prediction.
To make the resulting tensor optimization more tractable, $\ell_1$-norm surrogate is employed to relax $\ell_0$-norm and hence the resulting tensor optimization problem has a nonsmooth objective function due to the $\ell_1$-norm and orthogonal constraints from the orthogonal Tucker decomposition. To efficiently tackle this tensor optimization problem, we establish the first-order optimality condition in terms of stationary points, and then design a block coordinate descent (BCD) algorithm with convergence analysis and the computational complexity. Numerical results on BU-3DFE database and Bosphorus databases demonstrate the effectiveness of our proposed approach.


  %\medskip
  %\textbf{AMS subject classifications.}
\end{abstract}

\section{Introduction}
Facial expressions, as a way of no-verbal communication and a tool of delivering social information among human beings, are utilized to measure, interpretate and compute the human emotion. Hence, automatic recognition of facial expressions has attracted a great deal of interest, with wide applications in various domains such as psychology, human-machine interaction, transport security, health monitoring, computer graphics, pattern recognition, $etc$. And it  plays a crucial role in affective computing, computer vision and multimedia research \cite{COCE2016}.

\begin {figure*}[htb]
\centering%
\includegraphics[width=17cm]{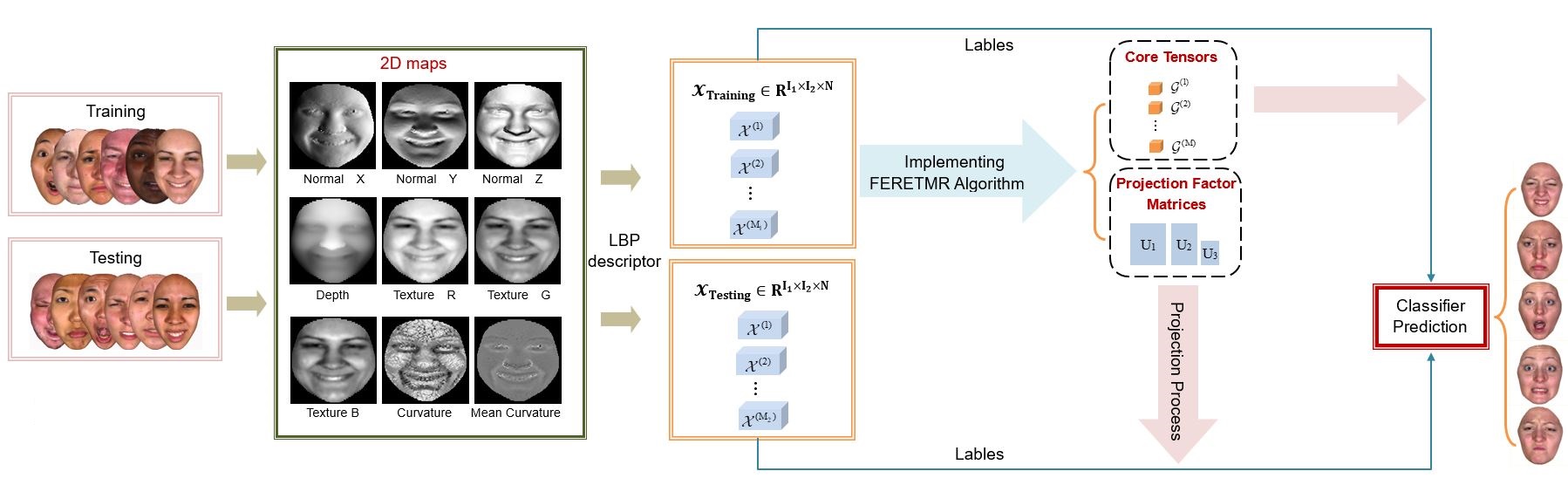}
\caption{A flowchart of the proposed approach (FERETMR) on BU-3DFE database.}
\label{chart}
\end {figure*}

In the last several decades, the majority of research for FER focused on 2D face images \cite{ ZLGS2004,WJYL2005} or video sequences \cite{ZHWC2016,DM2014} in terms of feature extraction based on facial expression, face detection and classification.  With the evolution of the algorithm in some close fields, such as face marking, feature engineering, its performance has been greatly improved.  However, 2D approaches based on the facial texture analysis have disadvantages suffering from illumination and pose variations, and possible occlusions \cite{PRLJ2000}.

With the fast development of 3D acquisition equipments, 3D FER by utilizing 3D face scans has gained a great deal of attention due to more robustness to illumination and pose variations.  Moreover, 3D face scans represented by 3D point sets have the ability to capture accurately the muscle movements of the facial skin surface, which are beneficial for FER.

The key of 3D facial expression recognition and analysis lies mainly on data description, feature extraction, and effective dimensionality reduction methods. As the very start in 3D FER, the data description is the most fundamental and crucial. A simple but widely used way of the data description in the literature on 3D FER is the {\it vectorization} (See, e.g., \cite{TH2008,ZHDC2010,Yurtkan2014Entropy, ZHWC2016, LDHWZ2015, JYJR2019, HSZC2017, YHWC2015,TZMYW2019,PRRMM2020}). However, the main drawback that vectorization suffers is the loss of the internal structure information of the data samples in which potential or inherent sparsity may hidden, and hence the dimensionality curse comes along by dismissing these favorable structural properties. To alleviate this issue, a more natural way to describe 3D facial expression data is using tensors, which not only maintains the spacial structure but also admits sparse representation when appropriate tensor decomposition is chosen, by employing tools from tensor analysis.
At present, the existing methods using tensors to describe 3D facial expression data are mostly based on tensor decomposition \cite{FRAJ2017,FRJA2019,FQLJAW2019,FQLAJ2021,FRJ2020,JRF2020,JR2021}, which have opened up a new technology direction and made some progress. However, the local structure (geometric information) of 3D tensor samples in these methods are not maintained in the low-dimensional tensors space during the dimensionality reduction.

To solve the above problem, a novel approach for 2D+3D FER via embedded tensor manifold regularization is proposed in this paper, with detailed flowchart shown in Fig.~\ref{chart}. At first,  a novel data representation is given, i.e., 3D tensors are constructed by stacking nine different features extracted from textured 3D scans to keep the structural information and correlations between multi-modal data (2D face images and 3D face shape models).  Tensor decomposition based on low-rank approximation is regarded as a powerful technique to be able to extract useful and discriminative low-dimensional information from the high-dimensional data.  And Tucker decomposition is one of the widely utilized forms of low-rank tensor decomposition, which decomposes a tensor into a product of a number of factor matrices and a core tensor.  We focus on Tucker decomposition in this paper. Based on orthogonal Tucker decomposition of the generated 3D tensors, our goal in this paper is to find projection factor matrices and a set of core tensors of relatively small sizes for the facial expression classification prediction.  Since potential similarities may inherit in the tensor modelling process, the resulting 3D tensors, probably of high dimension for real data sets, can be embedded into low-dimensional spaces.  A tensor reduction dimensionality technique is then utilized to the core tensors generated from orthogonal Tucker decomposition, equipped with an embedded tensor manifold regularization scheme to preserve the geometrical information during the dimension reduction.  Meanwhile, the $\ell_1$-regularization term is employed to promote sparsity structure on the involved core tensors.  Finally, an efficient optimization algorithm with the block coordinate descent (BCD) framework is designed to solve the resulting tensor optimization problem.  Thus, a novel tensor optimization model with an embedded tensor manifold regularization and sparsity based on the Tucker decomposition is then built to extract the useful and discriminative low-dimensional information from the generated 3D tensors model by using the tensor reduction dimensionality strategy. At the same time, Optimality analysis and stationarity are detailed according to the resulting optimization problem.  The analysis of convergence and the computational complexity of the proposed algorithm is also shown, where the computational complexity scales linearly with respect to the size of the constructed tensors and the number of the tensor samples, respectively.  To verify the effectiveness of our proposed approach, the multi-class-SVM is utilized for expression classification prediction.

The main contributions of our work are summarized below:
\begin{itemize}
	\item A novel data representation is given, in which a 3D tensor model is constructed by utilizing both 2D and 3D face data.  This kind of data representation overcomes the issues that the small sample size (SSS) problem and the dimensionality disaster due to vector representation.
	\item A tensor dimensionality reduction technique is utilized to the core tensors generated from orthogonal Tucker decomposition via an embedded tensor manifold regularization scheme to preserve the geometrical information during the dimensionality reduction.
	\item An efficient algorithm with the block coordinate descent (BCD) framework is designed to effectively solve the proposed tensor optimization model, and is applied into 2D+3D FER.
	\item  Optimality analysis and stationarity are detailed according to the resulting optimization problem. Meanwhile the convergence and computational complexity of the proposed approach are effectively analyzed.
\end{itemize}

 %\vskip 2mm
The rest of the paper is organized as follows. Related works are recalled including preliminaries on tensors in Section II.  The details of our proposed optimization model are  described in Section III. Experiment results and analysis are reported in Section IV, and conclusions are drawn in Section V.

\section{Related Works}
\subsection{Tensor Basics}\label{tensor-basic}
Throughout the paper, vectors will be written by lowercase letters, e.g., $x$, matrices by capital letters, e.g., $X$, and tensors by calligraphic letters, e.g., $\X$.  The symbols $\otimes$, $\circ$ and $\ast$ are used to denote the Kronecker, outer and Hadamard product, respectively.

Given an $N$th-order tensor $\X=\left(\X_{i_1\cdots i_N}\right)\in\R^{I_1\times I_2\ldots\times I_N}$, its mode-$n$ unfolding, denoted by $X_{(n)}$, is a matrix of size $I_n\times \prod_{k\neq {n},k=1}^N I_k$, with entries
$$\left(X_{(n)}\right){ij} = \X_{i_1\cdots i_N}, i = i_n, j = 1+\sum\limits_{k=1,k\neq n}^N(i_k-1)\prod_{m=1,m\neq n}^k I_m. $$
The mode-$n$ product of $\X$ with a matrix $U\in \R^{I_n\times R_n}$, termed as $\X\times_n U$, is a tensor $\Y\in\R^{I_1\times I_2\ldots \times R_n\times\ldots\times I_N}$ with its entries
$$\Y_{i_1 i_2\cdots r_n\cdots i_N} =\sum\limits_{i_n=1}^{I_n}\X_{i_1i_2\cdots i_n\cdots i_N} U_{r_n i_n}.$$
Given any two tensors $\X,~\Y\in \R^{I_1\times I_2\ldots\times I_N}$, the inner product $\langle\X, \Y\rangle$ is defined as the sum of all the products of their corresponding entries, that is,
$$\langle \X,\Y\rangle = \sum\limits_{i_1=1}^{I_1} \cdots \sum\limits_{i_N=1}^{I_N} \X_{i_1\cdots i_N}\Y_{i_1\cdots i_N}.$$
The tensor Frobenius norm, induced by the above inner product, is defined by $$\|\X\|_F:=\sqrt{\langle \X,\X\rangle}.$$
For sparsity characterization, the so-called $\ell_0$-norm (quasi-norm mathematically) of vectors can be naturally extended to high-order tensors, denoted by $\|\X\|_0$, which counts the number of nonzero entries in $\X$, i.e.,
$$\|\X\|_0 := \sharp \{(i_1, i_2, \cdots, i_N): \X_{i_1\cdots i_N}\neq 0\}.$$
Analogous to vectors, the tensor $\ell_1$-norm $\|\X\|_1$ also serves as the tighest convex surrogate of $\|\X\|_0$ and is defined by
$$\|X\|_1 := \sum\limits_{i_1=1}^{I_1}\sum\limits_{i_2=1}^{I_2}\ldots\sum\limits_{i_N=1}^{I_N}|\X_{i_1 i_2\ldots i_N}|,$$ respectively.

\subsection{Tensorial Data Reduction}\label{TDR}

\subsubsection{\bf Orthogonal Tucker Decomposition}
Tucker decomposition, which decomposes a tensor into a core tensor multiplied by a set of factor matrices along each mode, is one of the most widely used tensor decomposition methods.  Usually the orthogonality constraint is imposed to each factor matrix which yields the so-called orthogonal Tucker decomposition. Mathematically, for a given $N$th-order tensor $\X\in\R^{I_1\times I_2\ldots\times I_N}$, its orthogonal Tucker decomposition can be written as
$$ \X = \G\prod_{n=1}^N\times_n U_{n}, ~~\textrm{with}~ U_n\in \St(I_n,R_n), n = 1,\ldots, N,$$
where $\G\in\R^{R_1\times R_2\ldots\times R_N}$ is the core tensor, $\St(I_n,R_n):=\{U_n \in \R^{I_n\times R_n}|U_n^T U_n={\tt I}_{R_n}\}$ is the so-called the Stiefel manifold \cite{EAS1998}, and $U_n$'s are factor matrices which are partially orthogonal. Here ${\tt I}_{R_n}$ denotes the identity matrix of size $R_n\times R_n$.

\subsubsection{\bf Tensor Sparse Representation}
Sparse representation (SR) method, which is stemmed from compressed sensing (CS) \cite{DDL2006}, is widely applied into extensive application fields, such as pattern recognition, signal processing, machine learning, image processing, computer vision \cite{WMMSH2010,XZYY2011}, etc.  On the basis of the orthogonal Tucker decomposition, a structured sparse representation or approximation imposed on the core tensor can be obtained as follows
\begin{equation}\label{gs0}
\min\limits_{\{U_n\},\G} \left\{\|\G\|_0: \X = \G\prod_{n=1}^N\times_n U_{n}, ~U_n\in \St(I_n,R_n)\right\}.
\end{equation}
Since the $\ell_0$-norm is non-convex and discontinuous, problem (\ref{gs0}) is NP-hard generally and difficult to compute its exact optimal solutions.  One of the most popular relaxation strategies is to use the $\ell_1$-norm \cite{CH2015,ZXYLZ2017,Xu2015} as a surrogate since the $\ell_1$-norm is shown to be the tightest convex relaxation within the unit ball. The resulting continuous optimization problem takes the form of
\begin{equation}\label{g0}
\min\limits_{\{U_n\},\G} \left\{\|\G\|_1: \X = \G\prod_{n=1}^N\times_n U_{n}, ~U_n\in \St(I_n,R_n)\right\}.
\end{equation}

\subsubsection{\bf Manifold Regularization Extension}
Inspired by the work \cite{ZGCM2014,LNCYW2017,AZZJ2018} based on manifold learning, the manifold regularization framework is utilized to efficiently build a nearest neighborhood graph information for preserving the local geometry structure of higher order tensor data. The main idea is: given $M$ tensors of the same size $I_1\times I_2\cdots\times I_{N}$, say $\X^{(1)},\X^{(2)},\cdots,\X^{(M)}$, find partially orthogonal matrices $U_1$, $\cdots$, $U_N$ to decompose these $M$ tensors simultaneously in the Tucker sense, i.e.,
\begin{equation}\label{mani-reg}
\X^{(i)} = \G^{(i)} \times_1 U_1 \times_2 \cdots  \times_N U_N, ~~i\in [M],
\end{equation}
where $\G^{(i)}\in \R^{R_1\times \cdots \times R_N}$, $i=1,\cdots, M$ are of reduced dimensions, comparing to the original $\X^{(i)}$'s. Concisely, by stacking all $\X^{(i)}$'s and $\G^{(i)}$'s into $(N+1)$th-order tensors, namely $\X\in \R^{I_1\times \cdots \times I_N\times M}$ and $\G\in \R^{R_1\times \cdots \times R_N\times M}$, the above data reduction via decomposition can be written as
\begin{equation}\label{cons}
\X = \G\times_1 U_1\times_2 \cdots \times_N U_N.
\end{equation}
This is indeed an orthogonal Tucker decomposition for $\X$ with ${\tt I}_M$ as the last mode factor matrix. To preserve the potential or prior local similarity among these $M$ original tensors $\X^{(i)}$'s, one would expect to reflect such similarities among the resulting low-dimensional tensors $\G^{(i)}$'s. Adopt the following weight matrix $W\in \R^{M\times M}$ with entries
\[w_{ij}=\left\{\begin{array}{ll}
1,   &\text{if$~\X^{(i)}\in~\N_{\textrm{k}}(\X^{(j)})$~ or~$\X^{(j)}\in \N_{\textrm{k}}(\X^{(i)})$}\\
0, &\text{otherwise$$}.
\end{array}
\right.\]
where $\N_{\textrm{k}}(\cdot)$ is the set consisting of k-nearest neighbors of the object. To keep local similarity among low-dimensional projections  $\G^{(i)}$'s, it is then reasonable to construct the following Stiefel manifold constrained tensor optimization model
\begin{align}\label{mgr}
\min_{U_n \in\St(I_n,R_n)}\sum\limits_{i\neq j, i,j=1}^M\left\| \G^{(i)}-\G^{(j)}\right\|_F^2 w_{ij}.
\end{align}

\section{The Proposed Approach}

\subsection{The Manifold Regularization Orthogonal Tucker Decomposition Model}
\begin {figure}[H]
\centering%
\includegraphics[width=8.8cm,height=4.2cm]{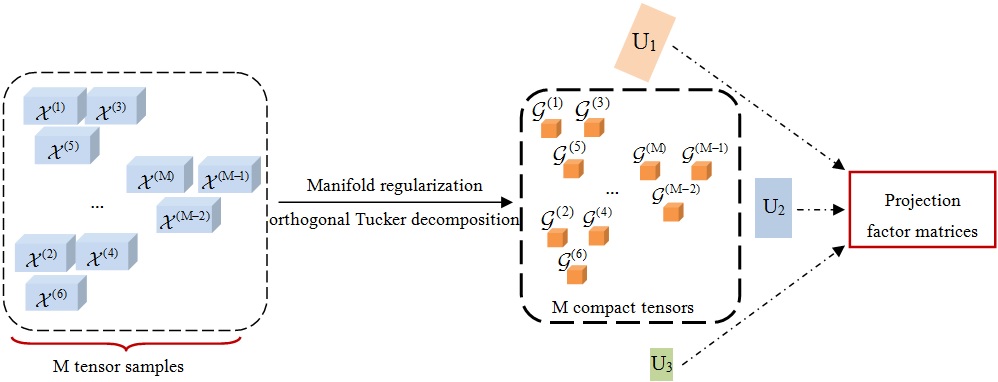}
\caption{Idea description of our proposed approach.}
\label{idea}
\end {figure}

Let $ \X^{(1)}, \cdots, \X^{(M)}\in {\R}^{I_1\times I_2\times N}$  be $M$ samples of generated 3D face data by naturally storing $N$ features with image size $I_1\times I_2$. Stacking all these samples into a $4$th-order tensor, denoted as $\X\in \R^{I_1\times I_2\times N\times M}$. Adopting the idea of sparse tensor representation, along with the low-rank Tucker decomposition with embedded tensor manifold regularization, as described in Subsection \ref{TDR}, we construct the following sparse tensor optimization for tensor data reduction
\begin{align}\label{op_model0}
\min~&  \L\left(\{\G^{(i)}\}_{i=1}^M,\{U_n\}_{n=1}^3\right) \nonumber\\
  {\rm s.t.~} & \G^{(i)}\in \R^{R_1\times R_2\times N},~i=1,\cdots,M,\nonumber\\
  & U_{n}\in \St(I_n,R_n), ~n =1, 2, 3.
\end{align}
Here the objective function is
\begin{align}
&\L\left(\{\G^{(i)}\}_{i=1}^M,\{U_n\}_{n=1}^3\right) \nonumber\\
= &\frac{1}{\gamma}\sum\limits_{i=1}^{M}\left\| \G^{(i)}\right\|_1 + \frac{1}{2}\sum\limits_{i=1}^{M}\left\| \X^{(i)}-\G^{(i)}\prod_{n=1}^3\times_n U_{n}\right\|_F^2\nonumber\\ &+\frac{1}{\beta} \sum\limits_{i\neq j, i,j \in [M]}\left\| \G^{(i)}-\G^{(j)}\right\|_F^2 w_{ij}
\end{align}
\normalsize
with the tradeoff parameters $\gamma$, $\beta>0$ for controlling the three terms corresponding to the sparsity of core tensors, the reconstruction error, and the tensor geometry information preservation, respectively, and $I_3 = R_3 = N$.

\subsection{Optimality Analysis and Stationarity}
Before the algorithm design for solving problem \eqref{op_model0}, we address the optimality analysis in terms of stationary points for theoretical preparation. Recall from \cite[Defintion 8.3]{VA} that the subdifferential of a proper closed function $f$ at a given point $x$ in its domain $\textrm{dom}f$, usually termed as $\partial f(x)$ is defined as
%\small
$$ \partial f(x): = \{v:\exists x^k\rightarrow x \textrm{~and~} v^k\in \hat{\partial}f(x^k), v^k\rightarrow v\}.$$
where
\begin{equation}
 \hat{\partial}f(x^k)  = \left\{v^k:\lim\limits_{y\rightarrow x^k, y\neq x^k}\frac{f(y)-f(x^k)-\langle v^k,y-x^k\rangle}{\|y-x^k\|}\geq 0\right\}. \nonumber
\end{equation}
%\normalsize
If $f$ is differentiable at $x$, then the subdifferential $\partial f(x)$ reduces to the gradient $\nabla f(x)$.

\begin{definition}\label{sta}
We call ${\H^*}:=\left(\{(\G^{(i)})^*\}_{i=1}^M,\{U^*_n\}_{n=1}^3\right)$ a stationary point of problem \eqref{op_model0} if $\O\in \partial F(\H^*)$, or equivalently,
\begin{equation}\label{stationary}
\left\{
  \begin{array}{ll}
    \O\in \partial_{\G^{(i)}} \L(\H^*), ~i=1,\cdots, M,& \hbox{ } \\
    O\in \nabla_{U_n} \L(\H^*)+N_{\St(I_n,R_n)}(U^*_n),~n=1,2,3, & \hbox{ }
  \end{array}
\right.
\end{equation}
where $N_{\St(I_n,R_n)}(U^*_n)$ is the normal space to $\St(I_n,R_n)$ at $U_n$, taking the form of
\[N_{\St(I_n,R_n)}(U^*_n) = \{U_n^*S: S\in \R^{R_n\times R_n}, S^T = S\}.\]
\end{definition}
The following theorem states the first-order optimality condition of problem \eqref{op_model0} in terms of the stationarity defined as above.
\begin{theorem}\label{first-order-opt}
For problem \eqref{op_model0}, if ${\H^*}=\left(\{(\G^{(i)})^*\}_{i=1}^M,\{U^*_n\}_{n=1}^3\right)$ is a local minimizer, then $\H^*$ is a stationary point, i.e., \eqref{stationary} holds at $\H^*$.
\end{theorem}

\subsection{Solving the Optimization Model}
We will adopt the block coordinate descent (BCD) scheme in the algorithm design for solving the proposed tensor optimization problem \eqref{op_model0}. By introducing the indicator function $\delta_{St(I_n, R_n)}$ defined as
\begin{equation}\label{indicator}  \delta_{\St(I_n, R_n)}(U) = \left\{
\begin{array}{ll}
0, & \hbox{if $U\in \St(I_n, R_n)$;} \\
+\infty, & \hbox{otherwise.}
\end{array}
\right.
\end{equation}
we can rewrite (\ref{op_model0}) into the following nonsmooth tensor optimization problem
\begin{equation}\label{op_model} \min\limits_{\{\G^{(i)}\}_{i=1}^M,\{U_n\}_{n=1}^3} \L\left(\{\G^{(i)}\}_{i=1}^M,\{U_n\}_{n=1}^3\right)+\sum\limits_{n=1}^3 \delta_{\St(I_n, R_n)}(U_n)
\end{equation}
Denote $F\left(\{\G^{(i)}\}_{i=1}^M,\{U_n\}_{n=1}^3\right)$  as the objective function in \eqref{op_model}. One can see that the involved nonsmooth terms in $F$ are separable in $\G^{(i)}$'s and $U_n$'s. This observation inspires us to employ the block coordinate descent (BCD) method (\cite{Tseng2001}) with the following updates in the Gauss-Siedel fashion:
\small
\begin{equation}\label{updates}
\left\{
\begin{array}{ll}
U_1^{[k+1]} = \arg\min\limits_{U_1} F\left( \{\G^{(i)}_{k}\}_{i=1}^M, U_1, U_2^{[k]}, U_3^{[k]} \right), & \hbox{} \\
U_2^{[k+1]} = \arg\min\limits_{U_2} F\left( \{\G^{(i)}_{k}\}_{i=1}^M, U_1^{[k+1]}, U_2, U_3^{[k]} \right), & \hbox{} \\
U_3^{[k+1]} = \arg\min\limits_{U_3} F\left( \{\G^{(i)}_{k}\}_{i=1}^M, U_1^{[k+1]}, U_2^{[k+1]}, U_3 \right), & \hbox{}  \\
\G^{(1)}_{k+1} = \arg\min\limits_{\G^{(1)}} F\left( \G^{(1)}, \G^{(2)}_k, \ldots, \G^{(M)}_k, \{U_n^{[k+1]}\}_{n=1}^3\right), & \hbox{} \\
\G^{(2)}_{k+1} = \arg\min\limits_{\G^{(2)}} F\left( \G^{(1)}_{k+1}, \G^{(2)}, \G^{(3)}_k, \ldots, \G^{(M)}_k, \{U_n^{[k+1]}\}_{n=1}^3\right), & \hbox{} \\
~~~~ \vdots  & \hbox{ } \\
\G^{(M)}_{k+1} = \arg\min\limits_{\G^{(M)}} F\left( \G^{(1)}_{k+1}, \ldots, \G^{(M-1)}_{k+1}, \G^{(3)}, \{U_n^{[k+1]}\}_{n=1}^3\right), & \hbox{}
\end{array}
\right.
\end{equation}
\normalsize
For simplicity, we will remove the iterate number $k$ and $k+1$ from all decision variables $\G^{(i)}$'s and $U_n$'s, and use $\hat{\G}^{(i)}$'s and $\hat{U}_n$'s for the new updates in the remainder of this subsection.

The subproblems for updating $U_n$'s take the form of
\begin{eqnarray}\label{Un_update}
\hat{U}_n &=& \arg\min\limits_{U_n} \frac{1}{2}\sum\limits_{i=1}^M \left\|X^{(i)}_{(n)}-U_n \Phi^i_{(n)}\right\|^2_F+ \delta_{\St(I_n,R_n)}(U_n) \nonumber\\
%&=& \arg\min\limits_{U_n}\left\langle U_n, -\sum\limits_{i=1}^M X^{(i)}_{(n)}{\Phi^i_{(n)}}^T\right\rangle + \delta_{\St(I_n,R_n)}(U_n)  \nonumber\\
&=&  \arg\max\limits_{U_n} \left\langle U_n, \sum\limits_{i=1}^M X^{(i)}_{(n)}{\Phi^i_{(n)}}^T\right\rangle - \delta_{\St(I_n,R_n)}(U_n)  \nonumber \nonumber\\
&=& Y_nZ_n^T=:\textrm{qf}\left(\sum\limits_{i=1}^M X^{(i)}_{(n)}{\Phi^i_{(n)}}^T\right),
\end{eqnarray}
where $\Phi_{n}^{(i)}=(\G^{(i)}\prod_{k=1,k\neq n}^3\times_k U_{k})_{(n)}$, $Y_n\in \St(I_n,R_n)$ and $Z_n\in \St(R_n,R_n)$ are the matrices consisting of the left- and right- singular vectors of the matrix argument, and qf$(\cdot)$ stands for the product matrix $Y_nZ_n^T$. Here the %$Y_n$ and $Z_n$ are the matrices consisting of the left- and right- singular vectors of the matrix $\sum\limits_{i=1}^M X^{(i)}_{(n)}{\Phi^i_{(n)}}^T$ in its SVD.
last equality is due to the \textit{von Neumann's trace inequality} in \cite{ML1975}.

\begin {figure}[!t]
\centering%
\includegraphics[width=4.5cm]{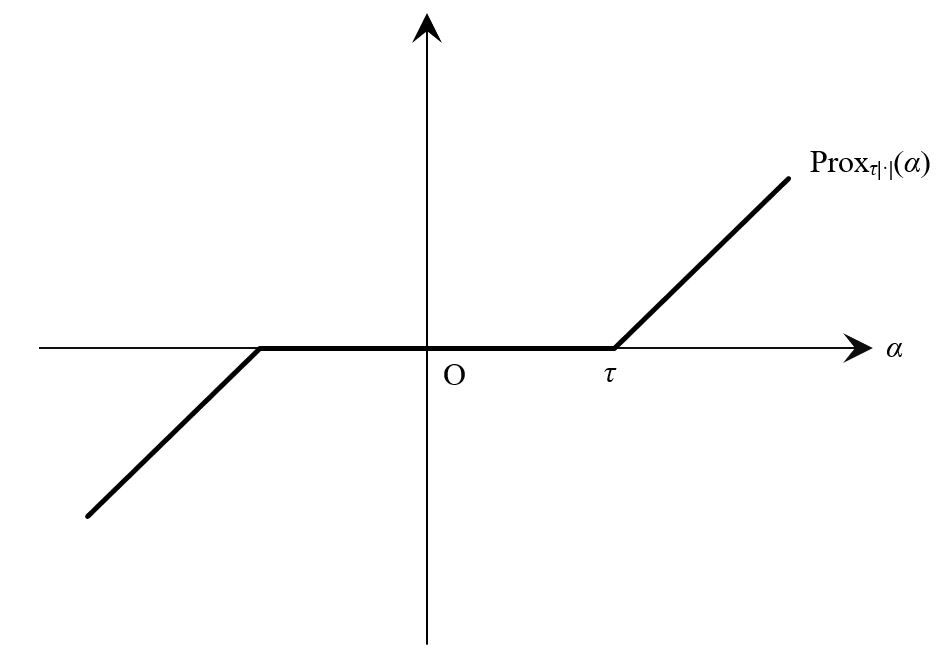}
\caption{The graph of Prox$_{\tau|\cdot|}(\alpha)$.}
\label{soft-tensor}
\end {figure}

The subproblems for updating $\G^{(i)}$'s take the form of
\small
\begin{eqnarray}\label{Gi_update0}
\hat{\G}^{(i)} &=& \arg\min\limits_{\G^{(i)}} \frac{1}{\gamma}\left\|\G^{(i)}\right\|_1 +\frac{1}{2}\left\|\X^{(i)}-\G^{(i)}\prod\limits_{n=1}^4 \times_n U_n\right\|^2_F\nonumber\\
&& ~~~~~~~+\frac{1}{\beta}\sum\limits_{j\neq i} \left\|\G^{(i)}-\G^{(j)} \right\|^2_F w_{ij} \nonumber\\
&=& \arg\min\limits_{\G^{(i)}} \frac{1}{\gamma}\left\|\G^{(i)}\right\|_1+\frac{1}{2}\left\|\X^{(i)}\prod\limits_{n=1}^3 \times_n U_n^T-\G^{(i)}\right\|^2_F \nonumber\\
&  &~~~~~~~+\frac{1}{\beta}\sum\limits_{j\neq i} \left\|\G^{(i)}-\G^{(j)} \right\|^2_F w_{ij} \nonumber\\
&=&\left(\arg\min\limits_{ \G^{(i)}_{i_1i_2i_3}} \frac{1}{\gamma}\left|\G^{(i)}_{i_1i_2i_3}\right|+\frac{1}{2}\left(\G^{(i)}_{i_1i_2i_3}-{\mathcal{D}}^{(i)}_{i_1i_2i_3}\right)^2 \right. \nonumber\\
& & \left.+\sum\limits_{j\neq i}\frac{w_{ij}}{\beta}\left(\G^{(i)}_{i_1i_2i_3}-\G^{(j)}_{i_1i_2i_3}\right)^2\right)_{i_1=1,i_2=1,i_3 =1}^{R_1,R_2,R_3}  \nonumber\\
&=& \left(\arg\min\limits_{ \G^{(i)}_{i_1i_2i_3}} \tau^{(i)}\left|\G^{(i)}_{i_1i_2i_3}\right|\right. \nonumber\\
& & \left.+\frac{1}{2}\left(\G^{(i)}_{i_1i_2i_3}-\alpha^{(i)}_{i_1i_2i_3}\right)^2\right)_{i_1=1,i_2=1,i_3 =1}^{R_1,R_2,R_3}    \nonumber\\
&=&\left(\text{Prox}_{\tau^{(i)}|\cdot|}(\alpha^{(i)}_{i_1i_2i_3})\right)_{i_1=1,i_2=1,i_3 =1}^{R_1,R_2,R_3},\nonumber\\
%&=&\nonumber
\end{eqnarray}
where $\mathcal{D}^{(i)} = \X^{(i)}\prod\limits_{n=1}^3 \times_n U_n^T$, $\tau^{(i)} = \frac{\beta}{\gamma\left(\beta+2\sum\limits_{j\neq i}w_{ij}\right)}$, $\alpha^{(i)}_{i_1i_2i_3} = \frac{\beta {\mathcal{D}}^{(i)}_{i_1i_2i_3}+\sum\limits_{j\neq i}w_{ij}\G^{(j)}_{i_1i_2i_3}}{\beta+2\sum\limits_{j\neq i}w_{ij}}$, and $\textrm{Prox}_{\tau^{(i)}|\cdot|}(\cdot)$
\normalsize
is the proximal mapping (See, e.g., \cite{VA}) associated with the function $f(t)=\tau^{(i)}|t|$, defined as
\begin{equation}\label{proximal}
\text{Prox}_f(x) = \arg\min\limits_{t} \left\{f(t)+\frac{1}{2}(t-x)^2\right\}.
\end{equation}
It is worth mentioning that the involved function $\textrm{Prox}_{\tau^{(i)}|\cdot|}(\cdot)$ for updating $\G^{(i)}$ has the following explicit formula

\begin{equation}\label{proximal1}
 \left(\text{Prox}_{\tau^{(i)}|\cdot|}(\alpha^{(i)}_{i_1i_2i_3})\right)_{i_1=1,i_2=1,i_3 =1}^{R_1,R_2,R_3}
 =\left(\max\{|\alpha^{(i)}_{i_1i_2i_3}|-\tau^{(i)},0\}\text{sign}(\alpha^{(i)}_{i_1i_2i_3})\right)_{i_1=1,i_2=1,i_3 =1}^{R_1,R_2,R_3}
\end{equation}

which is actually the tensor version of the so-called soft-thresholding operator (See, e.g., \cite{Beck2009}). For simplicity, we write the involved \textit{tensor soft-thresholding operator} as Prox$_{\tau^{(i)}\|\cdot\|_1}(\cdot)$. Thus, \eqref{Gi_update0} can be written as
\begin{equation}\label{Gi_update}
\hat{\G}^{(i)} = {\textrm{Prox}}_{\tau^{(i)}\|\cdot\|_1}\left(\frac{\beta \X^{(i)}\prod_{n=1}^3 \times_n U_n^T+ \sum\limits_{j\neq i}w_{ij}\G^{(j)}}{\beta+2\sum\limits_{j\neq i} w_{ij}}\right).
\end{equation} Such an operator is continuous, as illustrated via the graph of the univariate case shown in Fig. \ref{soft-tensor}.

\normalsize
The corresponding algorithmic framework is then presented in Algorithm \ref{IADM}.
\begin{algorithm}[!t]%[f1]
\caption{Solving Problem (\ref{op_model}) by BCD }
\label{IADM}
\begin{algorithmic}[2]
\Require
$M$ sample tensors $\{\X^{(i)}\}_{i=1}^M\in\mathbb {R}^{\mathnormal I_{1}\times I_{2}\times I_{3}}$; Parameters $\gamma$, $\beta$;
\Ensure
Factor matrices $\{U_n\}_{n=1}^3$;% A tensor $\G$ ;
\begin{itemize}
\item[Step~0] Randomly initialize $\{\G^{(i)}\}_{i=1}^M$, $\{U_n\}_{n=1}^3$; \nonumber
\item[Step~1] Update $U_n$'s and $\G^{(i)}$'s by (\ref{updates}) with closed form solutions stated in \eqref{Un_update} and \eqref{Gi_update};
\item[Step~2] If some stopping criterion is satisfied then stop, otherwise go to Step 1.
\end{itemize}
\end{algorithmic}
\end{algorithm}

\subsection{Computational Complexity and Convergence}
\subsubsection{Computational Complexity Analysis}
For each iteration, the computation cost for all $U_n$'s in \eqref{Un_update} is of the order
$$O(R_1R_2R_3(I_1+I_2+I_3))+O(MI_1I_2I_3(R_1+R_2+R_3)) +O(R_1^2I_1+R_2^2I_2+R_3^2I_3),$$
where the first term comes from the computation for all $\Phi_{(n)}^{(i)}$'s, the second term from $\sum\limits_{i=1}^M X^{(i)}_{(n)}{\Phi_{(n)}^{(i)}}^T$, and the last term from the SVD.

According to the closed-form solution in \eqref{Gi_update}, we can get the computation cost for updating all $\G^{(i)}$'s in each iteration of the order
$$O(MI_1I_2I_3(R_1+R_2+R_3))+O(\textrm{k}MR_1R_2R_3),$$
where the first term is the cost from the computation for all $\mathcal{D}^{(i)}$'s, and the second term for all involved $\tau^{(i)}$'s and all $\alpha_{i_1,i_2,i_3}^{(i)}$'s, in which $\textrm{k}$ represents the number of nearest neighbors employed in the mainifold construction.

Thus, the total cost in each iteration in Algorithm 1 is of the order
$$O(R_1R_2R_3(I_1+I_2+I_3+\textrm{k}M))+O(MI_1I_2I_3(R_1+R_2+R_3)) +O(R_1^2I_1+R_2^2I_2+R_3^2I_3).$$
As one can see, for each iteration, Algorithm 1 scales linearly with respect to the number
of tensor objects $M$, and also scales almost linearly with respect to the size of tensor objects $I_1I_2I_3$.

\subsubsection{Convergence Analysis}
As shown in Algorithm 1, the BCD scheme admits the non-increasing of the objective function $\cal{L}$ of our proposed tensor optimization problem \eqref{op_model0}, and the lower bound for increment from the current iterate to the next is estimated in the following theorem.

\begin{theorem}\label{convergence-thm}
Let $\left\{\left(\{\G_k^{(i)}\}_{i=1}^M,\{U^{[k]}_n\}_{n=1}^3\right)\right\}$ be the sequence generated by Algorithm {\bf 1}. Then the sequence of the objective values in problem \eqref{op_model0} is non-increasing, and for any given integer $k\geq 0$, we have
\begin{equation}\label{bound}
 \L\left(\{\G_k^{(i)}\}_{i=1}^M,\{U^{[k]}_n\}_{n=1}^3\right)-\L\left(\{\G_{k+1}^{(i)}\}_{i=1}^M,\{U^{[k+1]}_n\}_{n=1}^3\right)
\geq   \sum\limits_{i=1}^M \left(\frac{1}{2}+\sum\limits_{j\neq i}\frac{w_{ij}}{\beta}\right)\|\G_k^{(i)}-\G_{k+1}^{(i)}\|^2_F.
\end{equation}
\end{theorem}
The bound of the decrease of objective function values as stated in \eqref{bound} also indicates the stability of $\{\G_k^{(i)}\}$'s as $k$ grows. More specifically, we have $$\lim_{k\rightarrow \infty} \|\G^{(i)}_{k+1}-\G^{(i)}_{k}\|_F =0, ~~\forall i=1,\ldots,M,$$
since

\begin{eqnarray}
& &\sum\limits_{k=0}^{\infty} \sum\limits_{i=1}^M \left(\frac{1}{2}+\sum\limits_{j\neq i}\frac{w_{ij}}{\beta}\right)\|\G_k^{(i)}-\G_{k+1}^{(i)}\|^2_F \nonumber \\
 & \leq  & \sum\limits_{k=0}^{\infty}\left( \L\left(\{\G_k^{(i)}\}_{i=1}^M,\{U^{[k]}_n\}_{n=1}^3\right)-\L\left(\{\G_{k+1}^{(i)}\}_{i=1}^M,\{U^{[k+1]}_n\}_{n=1}^3\right)\right)\nonumber \\
 &\leq & \L\left(\{\G_0^{(i)}\}_{i=1}^M,\{U^{[0]}_n\}_{n=1}^3\right)  \nonumber
\end{eqnarray}
where the last inequality is due to the nonnegativity of the objective function $\L$.

Moreover, if the sequence generated by Algorithm \textbf{1} converges, it will converge to a stationary point of problem \eqref{op_model0}, as stated below.
\begin{theorem}\label{limiting-point}
If the sequence generated by Algorithm \textbf{1}, say $\left\{\H^k :=\left(\{\G^{(i)}_k\}_{i=1}^M,\{U^{[k]}_n\}_{n=1}^3\right)\right\}$, is convergent, i.e., there exists some $\H^*$ such that $\lim_{k\rightarrow \infty} \H^k = \H^*$, then $\H^*$ is a stationary point of problem \eqref{op_model0}.
\end{theorem}

\section{Experimental Evaluation}
In this section, numerical experiments will be implemented on the benchmark databases including the BU-3DFE database \cite{YWSWR2006} and the Bosphorus database \cite{SADCGS2008}, and comparison results to state-of-the-art methods in this domain and other Tucker decomposition based algorithms from other applications will be reported to evaluate the effectiveness of our proposed approach (FERETMR) in facial expression recognition.

\subsection{Implementation Details}

\subsubsection{Databases}
BU-3DFE database and the  Bosphorus database are two benchmark databases for FER. BU-3DFE database consists of 100 subjects (44 males and 56 females) with various ethnic backgrounds, and for each subject, there has a neutral expression and six prototypic expressions of four intensity levels, ranging from 1 to 4. The Bosphorus database contains 105 subjects composed of 65 males and 45 females in a variety of poses, expressions, and occlusion conditions.  Unlike BU-3DFE database, only 65 subjects have six facial expressions without intensity levels in Bosphorus database.

\subsubsection{Protocols}
Five different protocols, termed as Setups I to V, will be used in our numerical experiments, where Setup IV is tailored for the Bosphorus database and the other four are for BU-3DFE database. More specifically, 60 out of 100 subjects with samples of the two highest intensity levels, in a given fixed order and in random, are selected in Setups I and II respectively, and all these 100 subjects are selected in Setup III, for 10-fold cross-validation. For all these three setups, 100 rounds are conducted and the average score is obtained as the final recognition accuracy. To facilitate the comparison with other existing methods, a more flexible protocol Setup V is adopted with less than 20 times to calculate the final recognition accuracy (See the details in Table \ref{other_methods}). Since Bosphorus database only contains 65 subjects with six facial expressions, Setup IV uses the scheme in Setup II on BU-3DFE database, by replacing the total number 100 with 65.

\subsubsection{Feature Selection}
\begin {figure}[!t]
\centering%
\includegraphics[width=8cm]{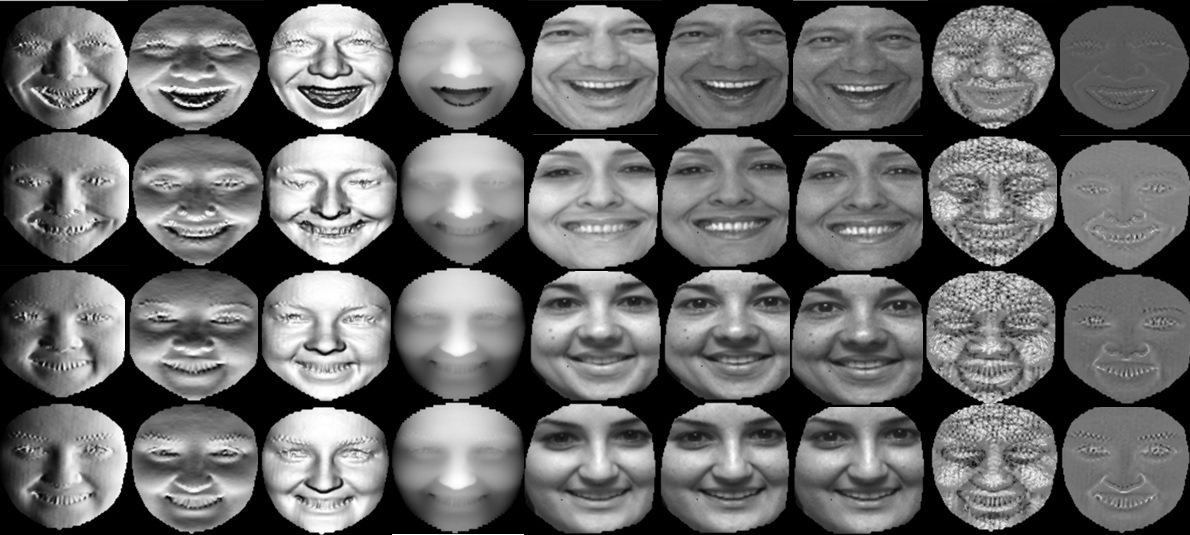}
\caption{\small{Illustration of different types of features of two 3D face scans with happiness expression in Bosphorus database. From top to bottom are subjects bs006, bs044, bs101 and bs104, and from left to right are Normal maps in the three directions (x, y, z), the depth maps, 3-channel 2D texture information (R, G, B), and curvature maps (curvature and mean curvature).}}
\label{bos_sample}
\end {figure}

In the experiments, we select nine effective and discriminative features to construct the 3D tensors $\{\X^{(i)}\}\in {\R}^{128\times 128\times 9}$ after LBP descriptor \cite{SGM2009} that is popularly applied into both 2D and 3D domain, the validation of which will be shown in Section IV(D).  In BU-3DFE database, these features by 2D maps are Depth map $I_g$, Normal maps in three directions $I_n^{x}, I_n^{y}$ and $I_n^{z}$, curvature maps (i.e., curvature $I_c$ and mean curvature $I_{mc}$), and Textured maps in three channels $I_t^r, I_t^g$ and $I_t^b$ that introduced in \cite{HSZC2017,LAMC2013}.  Different from BU-3DFE database,  the 3-channel textured features in Borsphorus database are obtained directly because the textured information of 3D face scans are provided poorly in Bosphorus database. It is worth mentioning that the 3-channel textured features in Bosphorus database are all masked by a common template to remove redundant parts. Fig. \ref{bos_sample} shows the nine types of features of 2D maps and 2D texture information of four 3D face scans in Bosphorus database.

\subsubsection{Parameter Setting}
To better tune the parameters $\gamma$, $\beta$ and $k$ (i.e., the $k$-nearest neighbors), we vary the values of $\gamma$ from $\{\textrm{1e}i\}_{i=1}^{10}$, $\beta$ from $\{\textrm{1e-}i\}_{i=1}^{10}$ and $k$ from $\{1,\ldots,10\}$, and find a best setting with $\gamma=$1e4, $\beta=$1e-6 and $k=4$.  The details about the parameter selection are shown in Section $D$.

\subsubsection{Stopping Criterion}
\begin{figure}[!t]
\centering
\subfigure[]{
 \label{RA}
  \includegraphics[width=8.0cm]{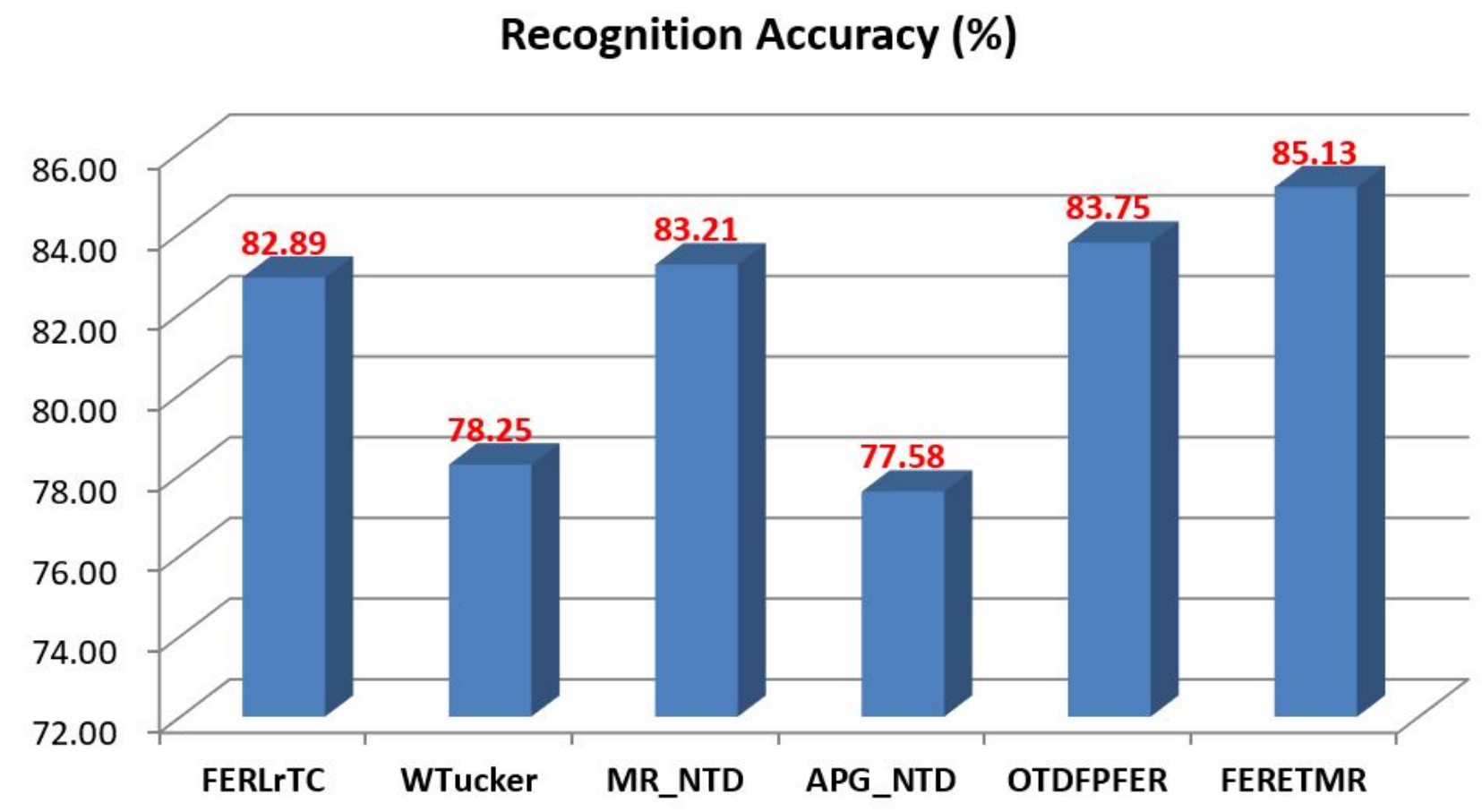}}
  \hspace{0.1in}
   \subfigure[]{
  \label{RE}
    \includegraphics[width=8.0cm]{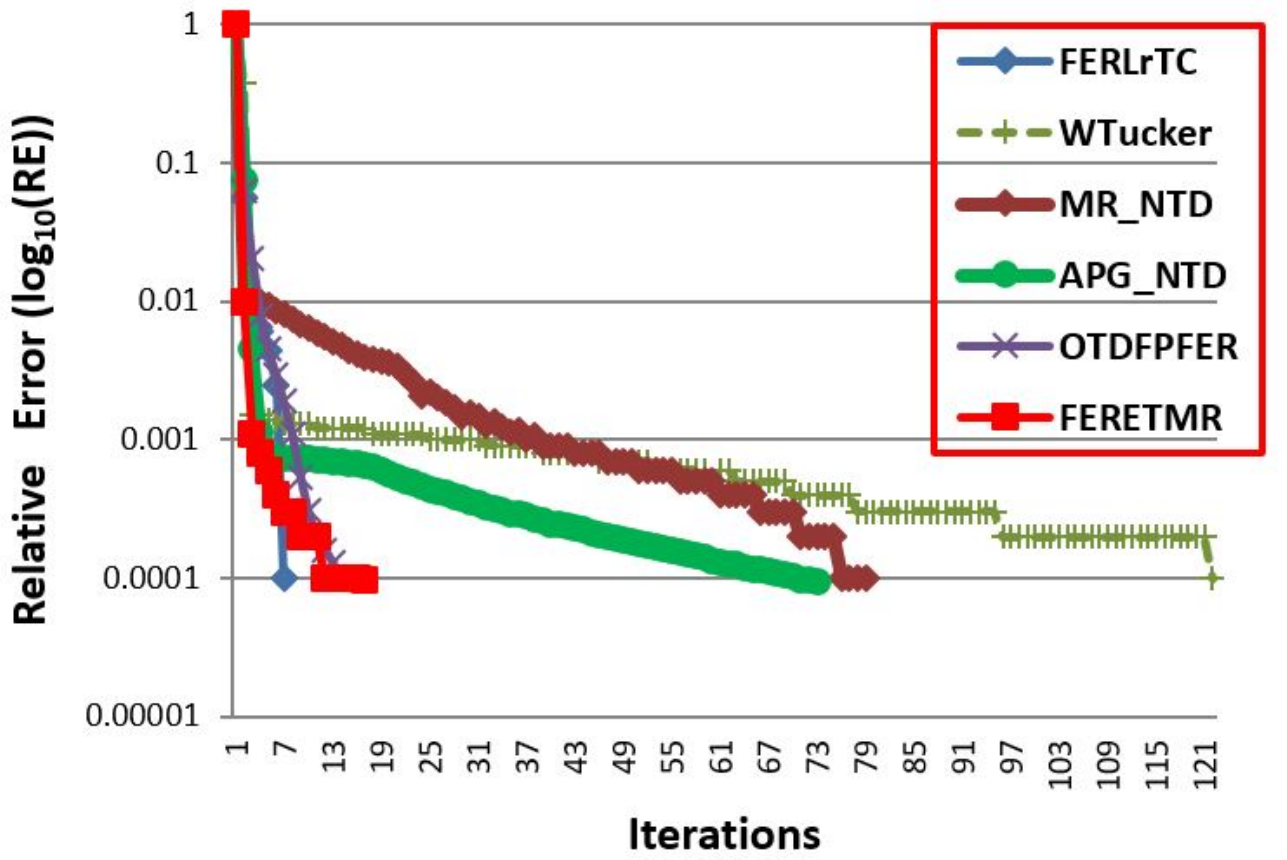}}
    \caption{Comparison results of average recognition accuracies and convergence behavior on BU-3DFE database by using Setup I.}
    \label{RERA}
    \end{figure}

As shown in Theorem \ref{convergence-thm}, our proposed approach generates a non-increasing sequence of objective values, which is also visualized in the red curve in Fig. \ref{RE}. Thus, it is natural to use the relative difference between two consecutive objective values for the stopping criterion of our algorithm:
\begin{align}
\frac{\left|\L(\{\G_{k+1}^{(i)}\}_{i=1}^M,\{U_n^{[k+1]}\}_{n=1}^3)-\L(\{\G_k^{(i)}\}_{i=1}^M,\{U_n^{[k]}\}_{n=1}^3)\right|}{\left\| \X\right\|_F}<\zeta\nonumber
\end{align}
where $\X=\lbrack\X^{(1)}, \X^{(2)}, \cdots \X^{(M)}\rbrack$ and $\zeta$ is the accuracy parameter which is set to be 1e-4 in our numerical experiments.

\subsection{Performance Evaluation on BU-3DFE Database}
Our first group of numerical experiments are implemented on BU-3DFE database. To evaluate the performance of our proposed approach, comparisons are conducted in three respects including the comparison by using our approach with Setups I, II and III as stated in Subsection A, the comparison of our approach with five other existing tensor Tucker decomposition based algorithms, and the comparison with more other state-of-the-art methods.

\subsubsection{Comparison in Different Protocols}
Table \ref{protocols} collects the comparison results on BU-3DFE database using Setups I, II and III.  As one can see, Setup I achieves the best result 85.13$\%$, which only uses the higher intensity levels, while Setup III obtains the worst 80.42$\%$, which utilizes four intensity levels from 1 to 4.  The comparison results indicates higher-intensity facial expressions are easier to be recognized than all-intensity expressions including lower-intensity levels (i.e., 1-level and 2-level).  Among the three protocols, two expressions of happiness and surprise are achieved better recognition results because of their higher facial deformation, whereas fear expression that can be confused with other five expressions is obtained worse results and is to a great degree confused with happiness expression.  Meanwhile, it can be found that sadness expression in Setup I is achieved the best recognition result among the three protocols, and even indicates a certain improvement compared with those in \cite{HSZC2017,YHWC2015,LAMC2013,FQLJAW2019,FQLAJ2021,JR2021}.
\begin{table}[!t]%*}[H]
\centering
\caption{Average confuse matrix on BU-3DFE database using different protocols.}
 \label{protocols}
 \scriptsize
\begin{tabular}{rrrrrrr}
\hline
\hline
  {\bf \%} &   {\bf AN} &   {\bf DI} &   {\bf FE} &   {\bf HA} &   {\bf SA} &   {\bf SU} \\
  \hline
   \textbf{AN} &     \textbf{ 78.85 } &      6.67  &      2.01  &      0.00  &     11.88  &      0.59  \\

 \textbf{DI} &      8.66  &   \textbf{   81.25}  &      5.52  &      2.61  &      0.00  &      1.96  \\

 \textbf{FE} &      1.89  &      8.23  &    \textbf{  76.51}  &      8.54  &      3.54  &      1.29  \\

 \textbf{HA}&      0.00  &      1.16  &      2.87  &      \textbf{95.53 } &      0.00  &      0.44  \\

  \textbf{SA} &     10.69  &      2.76  &      5.29  &      0.00  &    \textbf{  81.26}  &      0.00  \\

 \textbf{SU} &      0.17  &      0.33  &      1.58  &      0.56  &      0.00  &     \textbf{ 97.36}  \\

  \textbf{Setup I} & \textbf{85.13\%} &       &       &       &       &  \\

  {\bf \%} &   {\bf AN} &   {\bf DI} &   {\bf FE} &   {\bf HA} &   {\bf SA} &   {\bf SU} \\

  \textbf{AN} &    \textbf{ 79.12}  &      6.07  &      4.54  &      0.00  &      9.61  &      0.66  \\

  \textbf{DI}  &      7.22  &    \textbf{ 80.56}  &      5.04  &      1.65  &      2.44  &      3.09  \\

 \textbf{FE}  &      5.21  &      6.44  &     \textbf{71.49}  &      7.34  &      6.21  &      3.31  \\

 \textbf{HA} &      0.00  &      0.80  &      3.76  &     \textbf{95.03}  &      0.00  &      0.41  \\

  \textbf{SA} &     13.49  &      3.32  &      5.08  &      0.82  &     \textbf{77.29}  &      0.00  \\

 \textbf{SU}&      0.37  &      1.24  &      1.69  &      0.19  &      0.00  &     \textbf{96.51}  \\

\textbf{Setup II} & \multicolumn{6}{l}{\textbf{83.33\%}} \\

  {\bf \%} &   {\bf AN} &   {\bf DI} &   {\bf FE} &   {\bf HA} &   {\bf SA} &   {\bf SU} \\

    \textbf{AN}  &     74.76  &      7.97  &      3.39  &      0.00  &     12.99  &      0.89  \\

  \textbf{DI} &     10.28  &     77.99  &      4.65  &      1.16  &      3.17  &      2.75  \\

   \textbf{FE}  &      6.17  &      2.54  &     69.39  &     10.85  &      8.02  &      3.03  \\

   \textbf{HA} &      0.40  &      0.86  &      5.62  &     92.93  &      0.00  &      0.19  \\

   \textbf{SA} &     11.97  &      2.65  &     10.19  &      1.40  &     73.79  &      0.00  \\

  \textbf{SU} &      0.09  &      3.20  &      2.56  &      0.51  &      0.00  &     93.64  \\

 \textbf{Setup III} & \multicolumn{6}{l}{\textbf{80.42\%}} \\
\hline
\hline
\end{tabular}
\end{table}

\subsubsection{Comparison with Tucker Decomposition-based Algorithms}
Five state-of-the-art algorithms based on Tucker decomposition are compared with our proposed approach, which includes FERLrTC \cite{FQLJAW2019},  WTucker \cite{FMJA2015}, MR$\_$NTD \cite{LNCYW2017}, APG$\_$NTD \cite{Xu2015} and OTDFPFER \cite{FQLAJ2021}.
\begin{itemize}
\item FERLrTC:  A low-rank tensor completion approach via the nuclear-norm of factor matrices, together with the log-sum surrogate of the core tensor, is working on 4D tensor data, and a majorization minimization method is designed to solve the problem.
\item MR$\_$NTD: A manifold regularization term for the core tensors constructed in the Tucker decomposition is used to preserve geometric information in tensor data equipped with an alternating least squares manner.
\item APG$\_$NTD:  A sparse nonnegative Tucker decomposition approach is carried out via an alternating proximal gradient.
\item OTDFPFER:  An effective approach based on orthogonal Tucker decomposition using factor priors (OTDFPFER) is proposed to recognize 2D+3D facial expression automatically.
\item WTucker:  A low-rank tensor completion approach based on some presribed multilinear rank.
\end{itemize}

Among these approaches, OTDFPFER adopts the same cutting strategy as FERLrTC does. It is worth mentioning that the multilinear rank of APG$\_$NTD should be predefined, and the multilinear rank should be over-estimated in WTucker. Different from APG$\_$NTD, FERLrTC and OTDFPFER, our proposed approach use truncation accuracy thresholds to get $R_n$'s adaptively from tensor samples for dimensionality reduction, and MR$\_$NTD also adopts the same method as our proposed method to reduce the dimension.

The comparison results in terms of the average recognition accuracy and the convergence behavior are presented in Fig. \ref{RERA}. Specifically, Fig. \ref{RA} reports the comparison result of average recognition accuracies with Setup I, and Fig. \ref{RE} shows the comparison results of convergence behavior using the log relative error $\log_{10}$(RE), where $\textrm{RE}: = \|\X^{[k+1]}-\X^{[k]}\|_F/\|\X\|_F$
with $\X^{[k]}:=\left[\G^{(1)}_{k}\prod_{n=1}^3\times_n U_n^{[k]},\ldots, \G^{(M)}_{k}\prod_{n=1}^3\times_n U_n^{[k]}\right]$ and $\X = \left[\X^{(1)},\ldots, \X^{(M)}\right]$. As one sees, our approach outperforms the others in terms of both two important measures. Particularly, our approach needs less iterations then those in the algorithms which use the over-estimated or the predefined rank strategies, which shows the advantage of the low-rankness that we have utilized via dimension reduction.

\subsubsection{Comparison with Other Methods}
Our proposed approach is also compared with the state-of-the-art methods in the literature for FER, and the comparison results are listed in Table \ref{other_methods}, from which, one can see that our approach is quite competitive in all these four setups.
Note that he stability of the performance can not be guaranteed in the literatures \cite{WJYLW2006,SD2007,TH2008} that use the unstable protocol that run less than 20 times.  Overall, our proposed approach FERETMR obtains the better performance for 2D+3D FER under different protocols compared with the state-of-the-art methods.
\begin{table}[!t]%[!t]%*}
\scriptsize
\centering
\caption{Performance comparison with the state-of-the-art on BU-3DFE database (T shows the running times).}
 \label{other_methods}
 \begin{tabular}{c|c|c|c|c|c}
\hline
\hline
     Method &       Data &  Setup I  & Setup II  & Setup III & Setup V   \\
\hline
Yurtkan et al.  \cite{YKDH2014} &3D &   -       &    -        & - &88.28(8T)\\
Wang et al. \cite{WJYLW2006} &         3D &     61.79  &          - &-&83.60(20T) \\
Fu et al.  \cite{FRAJ2017} &         3D &     -       &    -        & -&85.802(10T)\\
Yurtkan et al.  \cite{Yurtkan2014Entropy}  &   3D   &    -        & - &     -&90.8(10T) \\

Tang et al. \cite{TH2008} &         3D &     74.51  &     -       &- &95.10(10T) \\

Soyel et al.  \cite{SD2007} &         3D &     67.52  &     -       &- &91.30(10T) \\
Lemaire et al.  \cite{LAMC2013} &         3D &     76.61  &     -   &     -  &-  \\
Gong et al.  \cite{GWLT2009} &         3D  &     76.22  &          - &      -&-    \\

Berretti et al.  \cite{BBPAD2010} &         3D  &          - &     77.54  &      -  &-   \\
Azazi et al. \cite{ALV2014}&         3D &     -       &    79.36        & -& -\\
Li et al.  \cite{LCHW2012} &         3D   &          - &     80.14  &    78.50   &-    \\
Zeng et al.  \cite{ZLCMG2013} &         3D  &      -   &   70.93  &    -    &-   \\

Fu et al.  \cite{FRJA2019} &         2D+3D&     82.36       &    81.78        & -& 95.12(10T) \\
%Walid et al. (2018) \cite{HTFBD2018}&         3D & landmark &         SVM &     -       &    -        & -&92.62(10T) \\
Zhao et al.  \cite{ZHDC2010} &      2D+3D&        -    &      -      & -&82.30(10T) \\
Fu et al.  \cite{FQLJAW2019} &      2D+3D & 82.89  & 80.91  & 78.96  &   95.28(10T)  \\
Yang et al.  \cite{YHWC2015} &         3D  &     84.80  &     82.73  &  -&        -  \\
Jiang et al.  \cite{JR2021} &      2D+3D & 83.31  & 80.75  & 74.1  &   -  \\
Fu et al.  \cite{FQLAJ2021} &      2D+3D & 83.75  & 81.63  & -  &   95.49(10T)  \\
ours &      2D+3D & 85.13  & 83.33  & 80.42  &   95.50(10T)  \\
\hline
\hline
\end{tabular}
\end{table}

There are also the state-of-the-art methods that go beyond our proposed approach with higher recognition rates, see, e.g., Table \ref{some_methods}.  It is worth mentioning that higher complexity is required to construct the networks in \cite{HSZC2017,CHWC2018}, or to get facial landmark localization in \cite{LDHWZ2015}, comparing to our approach.
\begin{table}[!t]
%\scriptsize
\centering
\caption{The state-of-the-art surppassing our proposed approach on BU-3DFE database with Setups I , II and III.}
 \label{some_methods}
 \newcommand{\tabincell}[2]
 {\begin{tabular}{@{}#1@{}}#2\end{tabular}}
\scriptsize
%\small
\begin{tabular}{c|c|c|c|c}

\hline
\hline
    Method &       Data &     Setup I (\%)  & Setup II (\%)& Setup III (\%)    \\
\hline
Chen et al. \cite{CHWC2018} &   3D   & 86.67 &85.96   & - \\
Li et al.  \cite{HSZC2017} &      2D+3D  & \tabincell{c}{86.86\\86.20} &          - & \tabincell{c}{81.04\\81.33}  \\
Li et al.  \cite{LDHWZ2015} &      2D+3D  &     86.32  &          - &      80.42  \\
\hline
\hline
\end{tabular}
\end{table}

\subsection{Performance Evaluation on Borphorus Database}
Table \ref{bos_protocol} reports the average confusion matrix by using Setup IV.  From this table, it is easily found that:  i) Happy and Sadness expressions obtain the highest and lowest recognition accuracies respectively, which means Happy expression is the easiest to be recognized, while sadness one is the most difficult; ii) Expressions with recognition accuracy less than 70$\%$ include digust, fear and sadness; iii) The confusion probability of angry expression with sadness expression is higher than others, and vice versa.  At the same time, the same is true of confusion of fear expression with surprise expression.  Compared with BU-3DFE database, Bosphorus database is very difficult to recognize facial expressions in this paper.
\begin{table}[!t]
\centering
\caption{Average confuse matrix on Bosphorus database using Setup IV.}
 \label{bos_protocol}
\scriptsize
\begin{tabular}{rrrrrrr}
\hline
\hline
  {\bf \%} &   {\bf AN} &   {\bf DI} &   {\bf FE} &   {\bf HA} &   {\bf SA} &   {\bf SU} \\
\hline
\textbf{ AN} &\textbf{ 75.29 } &      6.21  &      4.27  &      0.10  &     10.80  &      3.33    \\

\textbf{ DI} &      7.39  &\textbf{ 69.98 } &      6.32  &      5.79  &      8.11  &      2.41    \\

\textbf { FE} &      6.87  &      2.98  &\textbf{ 65.12 } &      2.36  &      4.37  &     18.30    \\

\textbf{ HA} &      0.00  &      3.37  &      2.29  &\textbf {93.56 } &      0.00  &      0.78    \\

\textbf{ SA} &     15.40  &     13.62  &      5.84  &      0.00  &\textbf{ 63.32 } &      1.82    \\

\textbf {SU} &      2.01  &      2.48  &      5.35  &      0.20  &      0.00  &\textbf {89.96 }   \\

 \textbf{Setup IV} & \multicolumn{6}{l}{\textbf{76.21\%}} \\

\hline
\hline
\end{tabular}
\end{table}

\subsubsection{Comparison with Tucker Decomposition-based Algorithms}
Like on BU-3DFE database, we use the same five algorithms based on Tucker decomposition (i.e., FERLrTC, WTucker, MR$\_$NTD, APG$\_$NTD and OTDFPFER) to compare with our proposed algorithm on Bosphorus database by using Setup IV.
\begin{figure}[H]%[!t]
\centering
\subfigure[]{
 \label{RA_bos}
  \includegraphics[width=8.0cm]{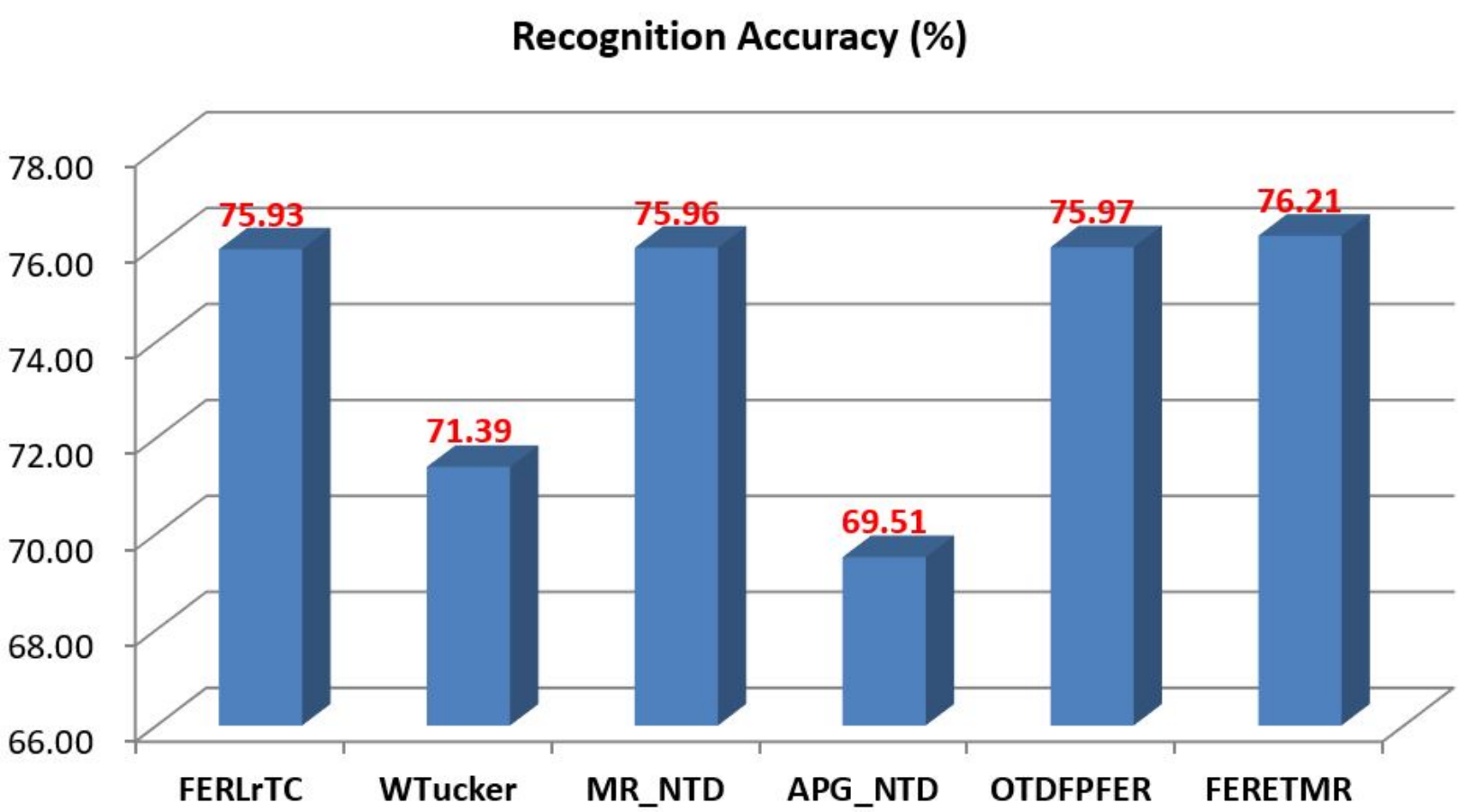}}%height=8cm,
  \hspace{0.1in}
   \subfigure[]{
  \label{RE_bos}
    \includegraphics[width=8.0cm]{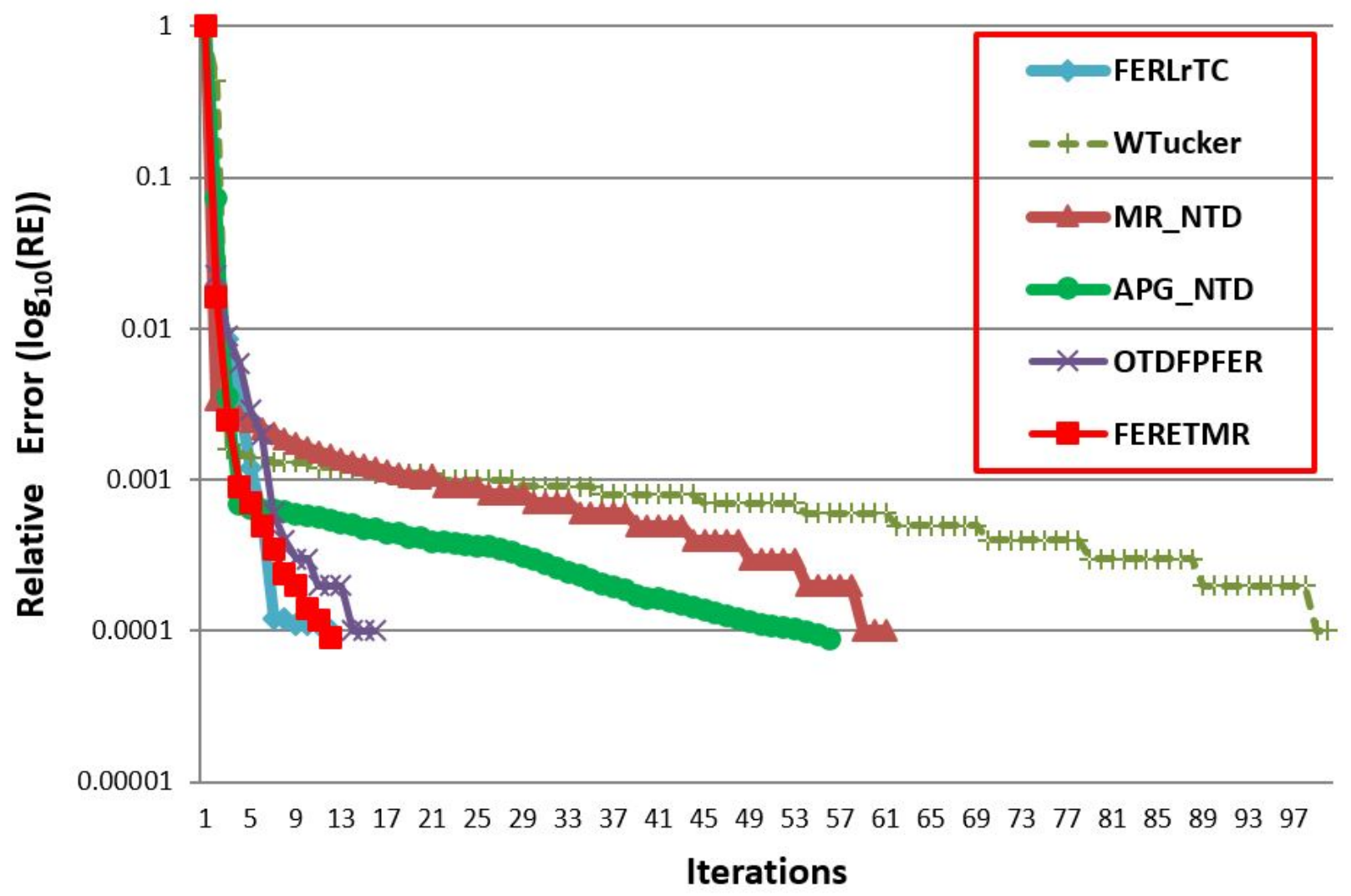}}%height=3.5cm,
    \caption{Comparisons of average recognition accuracies and convergence behavior on Bosphorus database using Setup IV.}
    \end{figure}

The comparison results of average recognition accuracies are indicated in Fig. \ref{RA_bos} by utilizing Setup IV.  As can be seen from this figure, our proposed method FERETMR achieves the best performance on recognition accuracy, whereas APG$\_$NTD obtains relatively the worst one.  The comparison results in Fig. \ref{RA_bos} fully illustrates that our proposed approach based on manifold regularization Tucker decomposition can extract effective features for 2D+3D facial expression.  Meanwhile it is easily found that our proposed approach has a great improvement compared with the recognition accuracies of FERLrTC and OTDFPFER.  These results demonstrate the Bosphorus database is very difficult to accomplish the task of 2D+3D facial expression recognition compared with BU-3DFE database. The comparison results of convergence behavior are reported in Fig. \ref{RE_bos} by the relative error on Bosphorus database with Setup IV.  Like on BU-3DFE database, the REs in this figure are also converge fast in limited number of iterations.  The results of analysis are the same as those on BU-3DFE database.

\subsubsection{Comparison with other Methods}
The performance comparisons with the state-of-the-art methods (i.e., \cite{DGADOB2018,LCHW2012,HS2014,FQLJAW2019,FQLAJ2021,JR2021}) are shown in Table \ref{bos_o} on Bosphorus database by using Setup IV.   From this table, we can observe that our proposed approach obtains the highest recognition accuracy, while the method \cite{HS2014} gains the lowest one.  Compared with the state-of-the-art methods in Table \ref{bos_o}, our proposed approach gains better performance on Bosphorus database by using Setup IV.
\begin{table}[!t]
\caption{Comparison with the state-of-the-art on Bosphorus database using Setup IV.}
 \label{bos_o}
 \centering
\begin{tabular}{c|c|c}
\hline
\hline
   Method &       Data &     Setup IV  \\
\hline
Ujir et al. \cite{HS2014} &  3D &       63.63 \\
Demisse et al. \cite{DGADOB2018} &      3D  & 67.05 \\
Li et al. \cite{LCHW2012} &  3D       &      75.83 \\
Fu et al. \cite{FQLJAW2019} &      2D+3D  & 75.93 \\
Fu et al. \cite{FQLAJ2021} &      2D+3D  & 75.97 \\
Jiang et al.  \cite{JR2021}&  2D+3D &       76.03 \\
Ours &      2D+3D  & {\bf 76.21} \\
\hline
\hline
\end{tabular}
\end{table}

\subsection{Discussion}

Four additional issues are discussed to further validate the effectiveness of our proposed approach (FERETMR).

\subsubsection{Parameter Selection}
To better obtain the recognition accuracies of facial expression on BU-3DFE and Bosphorus databases,  a best setting of the parameters $R_n$'s for dimensionality reduction is very important.  To determine the values of $R_n$'s, we follow the scheme in \cite{PACA2010}. Mathematically, for $n=1,2,3$,
$$R_n=\min\limits_{1\leq l\leq I_n}\left\{l: \frac{\sum_{j=1}^{l}\lambda_j}{\sum_{j=1}^{I_n}\lambda_j}\geq \sigma_n\right\},$$
where $\lambda_j$ is the $j$th largest eigenvalue of $\sum\limits_{i=1}^M X_{(n)}^{(i)}\left(X_{(n)}^{(i)}\right)^T$, and $\sigma_n$ is a threshold for truncation accuracy. Four different choices of $\sigma_n$'s are used in the experiments on BU-3DFE database with Setup I, and the recognition accuracy is reported in Table \ref{reduction}. As shown in Table \ref{reduction}, the second choice $(\sigma_1, \sigma_2, \sigma_3)= (0.90, 0.90, 0.9985)$ achieves the best accuracy among all the parameter settings. Thus, we simply use this setting in all experiments on both two databases.
\begin{table}[H]
 \caption{Comparison on average of recognition accuracy with different threshold fitness $\sigma$ in BU-3DFE database with Setup I.}
 \label{reduction}
\scriptsize
 \newcommand{\tabincell}[2]
{\begin{tabular}{@{}#1@{}}#2\end{tabular}}
\centering
\begin{tabular}{l|rrrr}
\hline
\hline
 Threshold fitness          &\tabincell{c}{$\sigma_1=0.85$,\\$\sigma_2$=0.85,\\$\sigma_3$=0.9985} &\tabincell{c}{$\sigma_1$=0.90,\\$\sigma_2$=0.90,\\$\sigma_3$=0.9985} &\tabincell{c}{$\sigma_1=0.92$,\\
$\sigma_2$=0.92,\\$\sigma_3$=0.9985} &\tabincell{c}{$\sigma_1$=0.95,\\$\sigma_2$=0.95,\\$\sigma_3$=0.9985} \\
\hline

 Accuracy (\%) &     81.75  &    {\bf 85.13 }  &  84.28 &     83.90  \\
\hline
\hline
\end{tabular}
\end{table}

\subsubsection{Combination Effectiveness with One Feature Excluded at One Time based on Feature-Level Fusion}
To verify the combination effectiveness of those nine features we have used, experiments on BU-3DFE and Bosphorus databases are conducted with one feature excluded at one time based on feature-level fusion.  Table \ref{fea_bu} reports the recognition accuracies and their differences with the average recognition accuracy $85.13\%$ on BU-3DFE database by utilizing Setup I for one feature excluded at one time.  From this table, we can easily observe that the difference is in the range $[-2.95, -1.01]$, among which the differences of $I_n^y$  and $I_{mc}^z$ achieve the lowest and highest, respectively.  For Bosphorus database, the results are shown in Table \ref{fea_bos}.  One can see that the differences with the average recognition accuracy $76.21\%$ are in the range $[-2.20,-0.48]$, and $I_g$ and $I_{mc}$ obtain the lowest and highest results, respectively.  The comparisons shown in Tables \ref{fea_bu}-\ref{fea_bos} demonstrate that there exists much complementarity between 2D and 3D modalities, and validate that any of nine kinds of features can not be excluded.  Therefore, it is effective to construct a set of 3D tensors by stacking nine types of features based on feature-level fusion.

\begin{table}[!t]
\caption{Comparisons of recognition accuracies for one feature excluded at one time.}
\centering
\subtable[On BU-3DFE database]{
\begin{tabular}{lrrrrr}
\hline
\hline
\% &          $I_n^x$ &          $I_n^y$ &          $I_n^z$ &          $I_t^r$&           $I_t^g$  \\
\hline
Setup I &     83.01  &     82.18  &     83.23  &     83.09  &     82.98  \\
Difference &     -2.12  &     -2.95  &     -1.90  &     -2.04  &     -2.15  \\
 \% &            $I_t^b$ &          $I_g$ &         $I_c$ &       $I_{mc}$  &  {\bf ALL} \\
 Setup I &     82.51  &     83.19  &     82.74  &     84.12  & \textbf{ 85.13 } \\
Difference &     -2.62  &     -1.94  &     -2.39  &     -1.01  &      0.00  \\
\hline
\hline
\end{tabular}
       \label{fea_bu}
}
\qquad
\subtable[On Bosphorus database]{
\begin{tabular}{lccccc}
\hline
\hline
\% &          $I_n^x$ &         $I_n^y$ &          $I_n^z$ &        $I_t^r$ &         $I_t^g$  \\
\hline
Setup IV &     74.25  &     74.12  &     74.52  &     74.22  &     75.07  \\
Difference &     -1.96  &     -2.09  &     -1.69  &     -1.99  &     -1.14  \\
\% &           $I_t^b$ &          $I_g$ &         $I_c$ &         $I_{mc}$  &  {\bf ALL} \\
 Setup IV &     74.90  &     74.01  &     74.66  &     75.73  & \textbf{ 76.21 } \\
Difference &     -1.31  &     -2.20  &     -1.55  &     -0.48  &      0.00  \\
\hline
\hline
\end{tabular}
       \label{fea_bos}
}
\end{table}

\subsubsection{Manifold Selection for Weight Strategy}
As introduced in Section II(A), a connection between a tensor $\X^{(i)}$ and another tensor $\X^{(j)}$ has the weight $w_{ij}$, and the matrix $W$ that is symmetric needs to be constructed.  In this paper, we only consider the following three ways: i) the binary weight strategy in which all the elements of the matrix $W$ are either 0 or 1; ii) the heat kernel weight scheme in which the weight value $w_{ij}$ is achieved by
$\exp(-\|\X^{(i)} - \X^{(j)}\|^2/\delta)$ where $\|\X^{(i)} - \X^{(j)}\|$ is the distance between $\X^{(i)}$ and $\X^{(j)}$.  Noted that the parameter $\delta$ is difficult to be determined. We set $\delta$ to be 2, 1000, 5000, respectively;  iii) the cosine weight strategy with $w_{ij}=\langle \X^{(i)}, \X^{(j)}\rangle/(\|\X^{(i)}\|\cdot\|\X^{(j)}\|$).
\begin{table}[!t]
\centering
%\tiny
\newcommand{\tabincell}[2]{\begin{tabular}{@{}#1@{}}#2\end{tabular}}
\caption{Comparisons of the recognition accuracies with different weight strategies on BU-3DFE database with Setup I.}
\begin{tabular}{l|ccccc}
\hline\hline
Weight Strategy &\tabincell{c}{Heat Kernel\\($\delta$=2)} & \tabincell{c}{Heat Kernel\\ ($\delta$=1000)} & \tabincell{c}{Heat Kernel\\ ($\delta$=5000)}  \\
Accuracy(\%) & 84.19 &      84.49 &      84.70   \\

Weight Strategy  &  Cosine & Binary \\
Accuracy(\%) &      84.32& {\bf 85.13}\\
\hline
\hline
\end{tabular}%
\label{bu_ws}%
\end{table}%

Table \ref{bu_ws} shows the comparison results with the three weight strategies on BU-3DFE database with Setup I.  From this figure, it is easily observed that the binary weight scheme achieves the best result, while the heat kernel weight strategy obtains the worst one when the parameter $\delta$ is set to 2.  Therefore, the binary weight strategy we use is more effective compared other schemes.

\subsubsection{Feature Descriptor Selection}
Since the recognition accuracy might be sensitive to the chosen local descriptor in feature extraction for accomplishing various tasks in 2D and 3D domains, we test the performances with four widely used descriptors including  HOG \cite{NDBT2005}, Dense-SIFT \cite{LYT2011}, Gabor \cite{ZLSA1998} and LBP on BU-3DFE database with Setup I. The comparisons of recognition accuracy  by using different descriptors are reported in Table \ref{descriptor}.
From this table, one can see that LBP gains the best result, while HOG obtains the worst one.  Meanwhile we can observe that Dense-SIFT and LBP represent better than HOG and Gabor, and the result of LBP is higher than those of HOG, Dense-SIFT, and Gabor by 6.26$\%$, 1.08$\%$, and 4.62$\%$, respectively.  Therefore, we adopt the LBP descriptor since it is effective and efficient to encode local structure of textons within an image patch (See also in \cite{YHYWC2018}).

\begin{table}[!t]
 \caption{Comparisons of recognition accuracy by using different feature descriptors on BU-3DFE database with Setup I.}
 \label{descriptor}
\centering
\begin{tabular}{c|ccccc}
\hline
\hline
Feature Descriptor & HOG & Dense-SIFT &  Gabor& LBP \\
\hline
Accuracy(\%) &  78.87 &      84.05 &      80.51 & {\bf 85.13} \\
 \hline
\hline
\end{tabular}
\end{table}

\section{Conclusion and Future Work}
In this paper, a 2D+3D facial expression recognition approach via embedded tensor manifold regularization (FERETMR) has been proposed and solved. By employing the $\ell_1$-norm and also the tensor manifold regularization among the core tensors of samples, together with the dimension reduction scheme via truncated low-rank Tucker decomposition, we have built a nonsmooth tensor optimization problem with Stiefel manifold constraints. The first-order optimality condition via stationarity has been established, and a BCD algorithm has been designed with analysis on theoretical convergence and computation complexity. Extensive numerical experiments have been conducted on BU-3DFE database and Bosphorus database which have illustrated the effectiveness of our proposed approach.

A possible future work will be exploiting other manifold information for our constructed 3D tensors, such as the optimal Laplacian matrix introduced in \cite{ZLYNXP2012} by considering both the local regression and global alignment.  Meanwhile it is necessary to properly extract more effective features from textured 3D face scans and establish a higher-order tensor model correspondingly.  The resulting tensor optimization will have a relatively large scale, and hence effective and robust algorithms are needed.  All of these will be the direction of our future efforts.

\section{Appendices}

\noindent{\bf Proof of Theorem \ref{first-order-opt}}
It is known from \cite[Theorem 10.1]{VA} that a local minimizer $\H^*$ of problem \eqref{op_model0} will satisfy $\O \in F(\H^*)$ and hence $\H^*$ is a stationary point of problem \eqref{op_model0} by Definition \ref{sta}. By employing the facts
$$\partial_{U_n} F(\H^*) = \nabla_{U_n} \L(\H^*)+\partial_{U_n} \delta_{\St(I_n,R_n)}(U_n)$$ and
\begin{equation}\label{stief-normal}
\partial_{U_n} \delta_{\St(I_n,R_n)}(U_n) = N_{\St(I_n,R_n)}(U_n)
\end{equation}
from \cite{VA} for $n=1,2,3$, we can derive that \eqref{stationary} holds at $\H^*$.

\noindent{\bf Proof of Theorem \ref{convergence-thm}}
The non-increasing property of the sequence of objective values follows readily from the BCD scheme. Specifically, the update scheme as presented in \eqref{updates} yields the following chain of inequalities
\small
\begin{eqnarray}\label{decreases}
 \L\left(\{\G_{k}^{(i)}\}_{i=1}^M,\{U_n^{[k]}\}_{n=1}^3\right)
 &\geq&\L\left(\{\G_{k}^{(i)}\}_{i=1}^M,U_1^{[k+1]},\{U_n^{[k]}\}_{n=2}^3\right)\nonumber\\
&\geq & \L\left(\{\G_{k}^{(i)}\}_{i=1}^M,\{U_n^{[k+1]}\}_{n=1}^2,U_3^{[k]}\right)\nonumber\\
&\geq & \L\left(\{\G_{k}^{(i)}\}_{i=1}^M,\{U_n^{[k+1]}\}_{n=1}^3\right)\nonumber\\
&\geq & \L\left(\G^{(1)}_{k+1},\{\G_{k}^{(i)}\}_{i=2}^M,\{U_n^{[k+1]}\}_{n=1}^3\right)\nonumber\\
&\vdots& \nonumber\\
&\geq & \L\left(\{\G_{k+1}^{(i)}\}_{i=1}^M,\{U_n^{[k+1]}\}_{n=1}^3\right).
\end{eqnarray}
\normalsize
To get the desired upper bound in \eqref{bound}, we define $$g_{\lambda,\alpha}(t):=\lambda|t|+\frac{\alpha}{2}(t-a)^2$$ with some $a\in \R$ and any given positive scalars $\lambda$ and $\alpha$. Denote
$$t^*:=\textrm{Prox}_{\frac{\lambda}{\alpha}|\cdot|}(a)=\arg\min\limits_{t\in \R} g_{\lambda,\alpha}(t).$$
From the optimality of $t^*$, we have
$$ 0\in \partial{g_{\lambda,\alpha,a}(t^*)}=\lambda\partial |t^*|+\alpha(t^*-a),$$
i.e., there exists some $\nu\in\partial |t^*|$, such that
\begin{align}\label{vt}
\lambda\nu+\alpha(t^*-a)=0.
\end{align}
Note that $|\cdot|$ is convex, and $\nu\in\partial |t^*|$. It follows from the definition of the subdifferential of convex functions that $|t|-|t^*|\geq\nu(t-t^*)$,$\forall t\in \R.$ Thus,
\begin{align}\label{gt}
&g_{\lambda,\alpha}(t)-g_{\lambda,\alpha}(t^*)\nonumber\\
=&\lambda(|t|-|t^*|)+\frac{\alpha}{2}((t-a)^2-(t^*-a)^2)\nonumber\\
\geq&\lambda\nu(t-t^*)+(\alpha(t^*-a)(t-t^*)+\frac{\alpha}{2}(t-t^*)^2)\nonumber\\
=&(\lambda\nu+\alpha(t^*-a))(t-t^*)+\frac{\alpha}{2}(t-t^*)^2\nonumber\\
=&\frac{\alpha}{2}(t-t^*)^2,
\end{align}
where the last equality is from \eqref{vt}.
Therefore,
\small
\begin{align}\label{coretensor}
&\L\left(\{\G_{k}^{(i)}\}_{i=1}^M,\{U_n^{[k+1]}\}_{n=1}^3\right)-\L\left(\G_{k+1}^{(1)},\{\G_{k}^{(i)}\}_{i=2}^M,\{U_n^{[k+1]}\}_{n=1}^3\right )\nonumber\\
&=\frac{1}{\gamma}\left(\left\|\G_{k}^{(1)}\right\|_1-\left\|\G_{k+1}^{(1)}\right\|_1\right)+
\frac{1}{2}\left\|\X^{(1)}-\G_k^{(1)}\prod\limits_{n=1}^3\times_n U_n^{[k+1]}\right\|_F^2\nonumber\\
&-\frac{1}{2}\left\|\X^{(1)}-\G_{k+1}^{(1)}\prod\limits_{n=1}^3\times_n U_n^{[k+1]}\right\|_F^2\nonumber\\
&+\frac{1}{\beta}\sum\limits_{j=2}^M\left(\left\|\G_{k}^{(1)}-\G_{k}^{(j)}\right\|_F^2-\left\|\G_{k+1}^{(1)}-\G_{k}^{(j)}\right\|_F^2\right)w_{1j}\nonumber\\
&=:\sum\limits_{i_1,i_2,i_3}
\left(g_{\frac{1}{\gamma},a^{(1)}}\left(\left(\G_{k}^{(1)}\right)_{i_1i_2i_3}\right)-g_{\frac{1}{\gamma},a^{(1)}}\left(\left(\G_{k+1}^{(1)}\right)_{i_1i_2i_3}\right)\right)\nonumber\\
&\geq\frac{a^{(1)}}{2}\sum\limits_{i_1,i_2,i_3}\left(\left(\G_{k}^{(1)}\right)_{i_1i_2i_3}-\left(\G_{k+1}^{(1)}\right)_{i_1i_2i_3}\right)^2\nonumber\\
&=\frac{a^{(1)}}{2}\left\|\G_{k}^{(1)}-\G_{k+1}^{(1)}\right\|_F^2,
\end{align}\normalsize
where $a^{(1)}:=1+2\sum\limits_{j\neq 1}^M \frac{w_{1j}}{\beta}$, and the last inequality is from \eqref{gt}.
\normalsize
Similarly, we have

\begin{equation}\label{coretensor2}
 \L\left(\G_{k+1}^{(1)},\{\G_{k}^{(i)}\}_{i=2}^M,\{U_n^{[k+1]}\}_{n=1}^3\right)
  -\L\left(\G_{k+1}^{(1)},\G_{k+1}^{(2)},\{\G_{k}^{(i)}\}_{i=3}^M,\{U_n^{[k+1]}\}_{n=1}^3\right )
 \geq\frac{a^{(2)}}{2}\left\|\G_{k}^{(2)}-\G_{k+1}^{(2)}\right\|_F^2, \end{equation}
$$~~~~~~~~~~\vdots  $$
\begin{equation} \L\left(\{\G_{k+1}^{(i)}\}_{i=1}^{M-1},\G_{k}^{(M)},\{U_n^{[k+1]}\}_{n=1}^3\right)-\L\left(\{\G_{k+1}^{(M)}\}_{i=1}^M,\{U_n^{[k+1]}\}_{n=1}^3\right ) \geq\frac{a^{(M)}}{2}\left\|\G_{k}^{(M)}-\G_{k+1}^{(M)}\right\|_F^2,
\end{equation}
with $a^{(i)}:=1+2\sum\limits_{j\neq i}\frac{w_{ij}}{\beta}, i=2,\dots,M.$
Consequently, for any given $k\geq 0$, we have

\begin{align}
&\L\left(\{\G_{k}^{(i)}\}_{i=1}^M,\{U_n^{[k]}\}_{n=1}^3\right)-\L\left(\{\G_{k+1}^{(i)}\}_{i=1}^M,\{U_n^{[k+1]}\}_{n=1}^3\right)\nonumber\\
\geq& \L\left(\{\G_{k}^{(i)}\}_{i=1}^M,\{U_n^{[k+1]}\}_{n=1}^3\right)-\L\left(\{\G_{k+1}^{(i)}\}_{i=1}^M,\{U_n^{[k+1]}\}_{n=1}^3\right )\nonumber\\
=&\L\left(\{\G_{k}^{(i)}\}_{i=1}^M,\{U_n^{[k+1]}\}_{n=1}^3\right)-\L\left(\G_{k+1}^{(1)},\{\G_{k}^{(i)}\}_{i=2}^M,\{U_n^{[k+1]}\}_{n=1}^3\right)\nonumber\\
&+\L\left(\G_{k+1}^{(1)},\{\G_{k}^{(i)}\}_{i=2}^M,\{U_n^{[k+1]}\}_{n=1}^3\right) -\L\left(\G_{k+1}^{(1)},\G_{k+1}^{(2)},\{\G_{k}^{(i)}\}_{i=3}^M,\{U_n^{[k+1]}\}_{n=1}^3\right)\nonumber\\
&~~~~~~~~~~\vdots   \nonumber\\
&+\L\left(\{\G_{k+1}^{(i)}\}_{i=1}^{M-1},\G_{k}^{(M)},\{U_n^{[k+1]}\}_{n=1}^3\right) -\L\left(\{\G_{k+1}^{(i)}\}_{i=1}^M,\{U_n^{[k+1]}\}_{n=1}^3\right)\nonumber\\
\geq & \sum\limits_{i=1}^M\frac{a^{(i)}}{2}\left\|\G_{k}^{(i)}-\G_{k+1}^{(i)}\right\|_F^2 = \sum\limits_{i=1}^M \left(\frac{1}{2}+\sum\limits_{j\neq i}\frac{w_{ij}}{\beta}\right)\|\G_k^{(i)}-\G_{k+1}^{(i)}\|^2_F,
\end{align}
\normalsize
where the first inequality follows from \eqref{decreases} and the last inequality from \eqref{coretensor} and \eqref{coretensor2}.

\noindent{\bf Proof of Theorem \ref{limiting-point}}
Several essential lemmas are proposed before proceeding the proof of Theorem \ref{limiting-point}.
\begin{lemma}\label{lem}
Let $\{A_k\}\subseteq \R^{m\times n}$ ($m\geq n$) be an infinite sequence and $A_k\rightarrow A_*$. Denote $B_k:=U_kV_k^T$ where $U_k \in \St(m,n)$ and $V_k\in \St(n\times n)$ are from the SVD $A_k = U_k \Sigma_k V_k^T$.  If $B_k\rightarrow B_*$, then there exist $U_* \in \St(m,n)$, $V_*\in \St(n\times n)$, and a diagonal matrix $\Sigma_*\in\R^{n\times n}$ such that $A_*=U_*\Sigma_*V_*^T$ and $B_* = U_*V_*^T$.
\end{lemma}
{\it Proof.} Since $\{V_k\}$ is bounded, there exists a convergent subsequence $V_{k_i}\rightarrow V_*$. It further yields that
$$ U_{k_i} = B_{k_i}V_{k_i}\rightarrow B_* V_*=:U_*,$$
which implies that $\Sigma_{k_i} = U_{k_i}^T A_{k_i}V_{k_i} \rightarrow U_*^T A_*V_*=:\Sigma_*.$
Thus, $B_{k_i} = U_{k_i}V_{k_i}^T\rightarrow U_*V_*^T=B_*$ since $B_k\rightarrow B_*$.
Similarly, we have $A_{k_i} = U_{k_i}\Sigma_{k_i}V_{k_i}^T\rightarrow U_* \Sigma_* V_*^T = A_*,$ since $A_k\rightarrow A_*$. This completes the proof.

By employing the optimality theorem for constrained nonlinear programming, we have the following lemma.
\begin{lemma}\label{lem2}
Given $X_1, \ldots, X_M\in \R^{m\times q}$, $\Phi_1, \ldots, \Phi_M\in \R^{n\times q}$, $U_*\in \R^{m\times n}$ $(q\geq m\geq n)$, let $A_*: = \sum\limits_{i=1}^M X_i\Phi_i^T$. If $U_*= \textrm{qf}(A_*)$, then
$$O\in \nabla\left(\frac{1}{2}\sum\limits_{i=1}^M\|X_i-U_*\Phi_i\|^2_F\right)+N_{\St(m,n)}(U_*).$$
\end{lemma}
{\it Proof.} Since $U_*= \textrm{qf}(A_*)$, we have
\begin{eqnarray}
U_* &=& \arg\max\limits_{U\in \St(m,n)} \left\langle A_*, U\right\rangle  \nonumber \\
&=& \arg\min\limits_{U\in \St(m,n)}\frac{1}{2}\sum\limits_{i=1}^M\|X_i-U\Phi_i\|^2_F \nonumber\\
&=& \arg\min\limits_{U\in \R^{m\times n}}\left\{\frac{1}{2}\sum\limits_{i=1}^M\|X_i-U\Phi_i\|^2_F+\delta_{\St(m,n)}(U)\right\}\nonumber
\end{eqnarray}
By invoking \cite[Theorem 10.1]{VA}, we have
$$O\in \partial \left(\frac{1}{2}\sum\limits_{i=1}^M\|X_i-U_*\Phi_i\|^2_F+\delta_{\St(m,n)}(U_*)\right).$$ Note that the function $\frac{1}{2}\sum\limits_{i=1}^M\|X_i-U\Phi_i\|^2_F$ is continuously differentiable in $U$. Combining with \eqref{stief-normal}, we have
\begin{align}
&\partial \left(\frac{1}{2}\sum\limits_{i=1}^M\|X_i-U_*\Phi_i\|^2_F+\delta_{\St(m,n)}(U_*)\right) \nonumber\\ =&\nabla\left(\frac{1}{2}\sum\limits_{i=1}^M\|X_i-U_*\Phi_i\|^2_F\right)+N_{\St(m,n)}(U_*).\nonumber
\end{align} This completes the proof of the lemma.

By utilizing the first-order optimality theorem for convex programming, we also have the following equivalence.
\begin{lemma}\label{lem3}
Given $\G^*$, $\W^*\in \R^{R_1\times R_2 \times R_3}$, and $\tau>0$,  then $\G^* = \textrm{Prox}_{\tau \|\cdot\|_1}(\W^*)$ if and only if
$$\O\in \partial (\tau \|\G^*\|_1)+\G^*-\W^*.$$
\end{lemma}

Now we are in a position to prove Theorem \ref{limiting-point}. According to the update scheme as presented in \eqref{updates}, together with the closed-form solutions established in \eqref{Un_update} and \eqref{Gi_update}, we have
\small
\begin{equation}\label{closed-eq}
\left\{
  \begin{array}{ll}
    U_1^{[k+1]} = \textrm{qf}\left(\sum\limits_{i=1}^M X^{(i)}_{(1)}\left(\G^{(i)}_k\times_2 U_2^{[k]}\times_3 U_3^{[k]}\right)_{(1)}\right), & \hbox{ } \\
    U_2^{[k+1]} = \textrm{qf}\left(\sum\limits_{i=1}^M X^{(i)}_{(2)}\left(\G^{(i)}_k\times_1 U_1^{[k+1]}\times_3 U_3^{[k]}\right)_{(2)}\right), & \hbox{ } \\
    U_3^{[k+1]} = \textrm{qf}\left(\sum\limits_{i=1}^M X^{(i)}_{(3)}\left(\G^{(i)}_k\times_1 U_1^{[k+1]}\times_2 U_2^{[k+1]}\right)_{(3)}\right), & \hbox{ } \\
     \G^{(1)}_{k+1} = {\textrm{Prox}}_{\tau^{(1)}\|\cdot\|_1}\left(\frac{\beta \X^{(1)}\prod_{n=1}^3 \times_n (U^{[k+1]}_n)^T+ \sum\limits_{j\neq 1}w_{1j}\G_k^{(j)}}{\beta+2\sum\limits_{j\neq 1} w_{1j}}\right), & \hbox{ } \\
    \vdots  & \hbox{ } \\
    \G^{(M)}_{k+1} = {\textrm{Prox}}_{\tau^{(M)}\|\cdot\|_1}\left(\frac{\beta \X^{(M)}\prod_{n=1}^3 \times_n (U^{[k+1]}_n)^T+ \sum\limits_{j\neq M}w_{Mj}\G_{k+1}^{(j)}}{\beta+2\sum\limits_{j\neq M} w_{Mj}}\right), & \hbox{ }
  \end{array}
\right.
\end{equation}
\normalsize
From the hypothesis, we have
\begin{equation}
\left\{
  \begin{array}{ll}
    \lim_{k\rightarrow \infty} \G^{(i)}_k = \left(\G^{(i)}\right)^*,~i=1,\ldots, M, & \hbox{ } \\
    \lim_{k\rightarrow \infty} U_n^{[k]} = U_n^*, ~n=1,2,3.& \hbox{}
  \end{array}
\right.
\end{equation}
Taking limit $k\rightarrow \infty$ on both sides of all equations in \eqref{closed-eq}, it then follows from Lemma \ref{lem} and the continuity of Prox$_{\tau^{(i)}\|\cdot\|_1}(\cdot)$ (See, Fig. \ref{soft-tensor}) that
\small
\begin{equation}
\left\{
  \begin{array}{ll}
    U_1^* = \textrm{qf}\left(\sum\limits_{i=1}^M X^{(i)}_{(1)}\left(\left(\G^{(i)}\right)^*\times_2 U_2^*\times_3 U_3^*\right)_{(1)}\right), & \hbox{ } \\
    U_2^* = \textrm{qf}\left(\sum\limits_{i=1}^M X^{(i)}_{(2)}\left(\left(\G^{(i)}\right)^*\times_1 U_1^*\times_3 U_3^*\right)_{(2)}\right), & \hbox{ } \\
    U_3^* = \textrm{qf}\left(\sum\limits_{i=1}^M X^{(i)}_{(3)}\left(\left(\G^{(i)}\right)^*\times_1 U_1^*\times_2 U_2^*\right)_{(3)}\right), & \hbox{ } \\
     \left(\G^{(1)}\right)^* = {\textrm{Prox}}_{\tau^{(1)}\|\cdot\|_1}\left(\frac{\beta \X^{(1)}\prod_{n=1}^3 \times_n U^*_n+ \sum\limits_{j\neq 1}w_{1j}\left(\G^{(j)}\right)^*}{\beta+2\sum\limits_{j\neq 1} w_{1j}}\right), & \hbox{ } \\
    \vdots  & \hbox{ } \\
    \left(\G^{(M)}\right)^* = {\textrm{Prox}}_{\tau^{(M)}\|\cdot\|_1}\left(\frac{\beta \X^{(M)}\prod_{n=1}^3 \times_n U^*_n+ \sum\limits_{j\neq M}w_{Mj}\left(\G^{(j)}\right)^*}{\beta+2\sum\limits_{j\neq M} w_{Mj}}\right), & \hbox{ }
  \end{array}
\right.
\end{equation}
\normalsize
In virtue of Lemmas \ref{lem2} and \ref{lem3}, we can conclude that $\H^* =\left(\{\left(\G^{(i)}\right)^*\}_{i=1}^M, \{U_n^*\}_{n=1}^3\right)$ satisfies \eqref{stationary} and hence $\H^*$ is a stationary point of problem \eqref{op_model0}.

\section*{Acknowledgment}
The authors sincerely appreciated the editor and anonymous referees for their valuable comments and suggestions. This work was partly supported by the National Natural Science Foundation of China (61471032, 11771038, 61772067), the fundamental research funds for the central universities (2017JBZ108), Beijing Natural Science Foundation (Z190002), Innovation Capability Improvement Plan Project of Hebei Science and Technology Department(21557611K), and Hebei Province Internet of Things Intelligent Perception and Application Technology Innovation Center.

%\bibliographystyle{plain}
%\bibliography{final211}

\end{document}